%% file: main.tex
\documentclass[10pt,journal,compsoc]{static/package/IEEEtran}
\usepackage{diagbox,amsmath,amsthm,amssymb,booktabs,comment,graphicx,subfigure,multirow,bm,tikz}
\usepackage{tabularx,array,float,colortbl,lstautogobble,color,zi4,listings,bbm}
\usepackage[ruled,linesnumbered]{algorithm2e}
\usepackage[symbol]{footmisc}
\usepackage{soul}
\usepackage{color,xcolor}
\usepackage{setspace,hyperref}
\usepackage{lineno}
\theoremstyle{definition}
\newtheorem{definition}{Definition}[section]

\newcommand{\tabincell}[2]{\begin{tabular}{@{}#1@{}}#2\end{tabular}}

\soulregister{\cite}7 
\soulregister{\citep}7 
\soulregister{\citet}7 
\soulregister{\ref}7 
\soulregister{\pageref}7 
\ifCLASSOPTIONcompsoc
  \usepackage[nocompress]{cite}
\else
  \usepackage{cite}
\fi
\newcommand{\partitle}[1]{\smallskip \noindent \textbf{#1.}}
\begin{document}
\title{\textsc{RemovalNet}: DNN Fingerprint Removal Attacks}
\author{
  Hongwei~Yao,
  Zheng~Li,
  Kunzhe~Huang,
  Jian~Lou,
  Zhan~Qin~$^{\dagger}$,
  and Kui~Ren,~\IEEEmembership{Fellow,~IEEE}
  \IEEEcompsocitemizethanks{
    \IEEEcompsocthanksitem Hongwei~Yao, Kunzhe~Huang, Zhan~Qin, and Kui~Ren
    are with the College of Computer Science and Technology, Zhejiang University, Hangzhou 310027, China. E-mail:\{yhongwei,hkunzhe,qinzhan,kuiren@zju.edu.cn\}.
    \IEEEcompsocthanksitem Hongwei~Yao, Jian~Lou, Kunzhe~Huang, Zhan~Qin, and Kui~Ren are with ZJU-Hangzhou Global Scientific and Technological Innovation Center, Hangzhou, China.
    \IEEEcompsocthanksitem Zheng~Li is with the German National Big Science Institution within the Helmholtz Association, Saarbrücken, German, E-mail:zheng.li@cispa.de.
    \IEEEcompsocthanksitem Zhan~Qin is corresponding author. This work is supported by the National Key Research and Development Program of China under Grant 2020AAA0107705, and the National Natural Science Foundation of China under Grant U20A20178 and 62072395.
  }
}

\markboth{IEEE TRANSACTIONS ON DEPENDABLE AND SECURE COMPUTING, VOL. 1, NO. 1, APRIL 2023}%
{Shell \MakeLowercase{\textit{et al.}}: Bare Advanced Demo of IEEEtran.cls for IEEE Computer Society Journals}

\IEEEtitleabstractindextext{
  \begin{abstract}
  With the performance of deep neural networks (DNNs) remarkably improving, DNNs have been widely used in many areas.
  Consequently, the DNN model has become a valuable asset, and its intellectual property is safeguarded by ownership verification techniques (e.g., DNN fingerprinting).
  However, the feasibility of the DNN fingerprint removal attack and its potential influence remains an open problem.
  In this paper, we perform the first comprehensive investigation of DNN fingerprint removal attacks. 
  Generally, the knowledge contained in a DNN model can be categorized into general semantic and fingerprint-specific knowledge.
  To this end, we propose a min-max bilevel optimization-based DNN fingerprint removal attack named \textsc{RemovalNet}, 
      to evade model ownership verification. 
  The lower-level optimization is designed to remove fingerprint-specific knowledge. 
  While in the upper-level optimization, we distill the victim model’s general semantic knowledge to maintain the surrogate model’s performance.
  We conduct extensive experiments to evaluate the \textbf{fidelity}, \textbf{effectiveness}, and \textbf{efficiency} 
      of the \textsc{RemovalNet} against four advanced defense methods on six metrics.
  The empirical results demonstrate that 
      (1) the \textsc{RemovalNet} is \textbf{effective}. 
          After our DNN fingerprint removal attack, the model distance between the target and surrogate models 
          is {$\times100$ times} higher than that of the baseline attacks, 
      (2) the \textsc{RemovalNet} is \textbf{efficient}. 
          It uses only {0.2\% (400 samples)} of the substitute dataset and 1,000 iterations to conduct our attack.
          Besides, compared with advanced model stealing attacks, the \textsc{RemovalNet} {saves nearly $85\%$} of computational resources at most,
      (3) the \textsc{RemovalNet} achieves high \textbf{fidelity} 
          that the created surrogate model maintains high accuracy after the DNN fingerprint removal process.
  \end{abstract}
  \begin{IEEEkeywords}
    DNN fingerprint removal, DNN fingerprinting, ownership verification.
  \end{IEEEkeywords}
}
\maketitle
\IEEEdisplaynontitleabstractindextext
\IEEEpeerreviewmaketitle

\section{Introduction}
\IEEEPARstart{A}{s} the performance of deep neural networks (DNNs) has remarkably improved, DNNs have been widely used in many areas 
    (e.g., image recognition~\cite{simonyan2014very,brock2021high} 
    and natural language processing~\cite{Young20ERASER,bragg2021flex}).
However, training a high-performance DNN model requires 
    tremendous training data and computational resources.
Especially, confidential datasets (e.g., medical, biological data, and facial images) 
    are collected and protected by organizations that are reluctant to open-source.
In this practical context, the DNN model becomes a valuable asset, which prompts the adversary to steal the victim model instead of constructing a model from scratch
    ~\cite{shi2019rapid,wu2019heterogeneous,shao2021towards}.

Recently, many ownership verification techniques have been proposed to protect 
    the intellectual property (IP) of the DNN models.
Given a victim model and a suspected model, the ownership verification algorithm aims to determine 
    whether the suspected model is derived from the victim model. 
Depending on the verification strategy, the DNN ownership verification algorithms can be broadly categorized into watermarking and fingerprinting.
DNN watermarking ~\cite{uchida2017embedding,zhang2018protecting,adi2018turning,
    darvish2019deepsigns,masoumeh21robustness,lukas2022sok}
    is an active ownership verification technology 
    that actively embeds an imperceptible watermark (e.g., noise or trigger pattern) into the redundant weights of the victim model.
During the evaluation phase, the stealing action can be detected 
    if a similar watermark can be extracted from the suspected model.
Different from watermarking, DNN fingerprinting
    ~\cite{wang2021fingerprinting, peng2022fingerprinting, cao2021ipguard, pan2022metav, chen2022copy,jia2021zest,lukas2021deep, li2021modeldiff, yang2022metafinger}
    is a passive ownership verification technology.
DNN fingerprinting is non-invasive in 
    that it only extracts a sequence of probing samples (i.e., fingerprints) 
    from the victim model without modifying the model itself.
During the verification phase, 
    the verifier traces the behavioral patterns of the DNN model using the extracted probing samples.
DNN fingerprinting leaves the victim model unaltered, whereas watermarking involves modifying it.

On the other hand, consider a \textit{post-stealing} scenario, i.e., ownership verification attacks.
After stealing from the victim model, the adversary utilizes the victim model to derive a surrogate model, attempting to bypass ownership verification systems.
In most literature of current studies, the attackers focus on detecting and removing watermarks on the DNN model
    ~\cite{cun2021split, chen2021refit, wang2021riga, qi0022attention}.
However, the feasibility of DNN fingerprint removal is still an open problem, and understanding whether DNN fingerprinting can suffer from such removal attacks is a pressing question.
DNN fingerprint removal attacks undermine the confidentiality and reliability of ownership verification algorithms, causing a loss of trust and confidence in the system. 
Removing the DNN fingerprint from the victim model while maintaining its performance involves three key challenges.
Firstly, fingerprints are intrinsic characteristics of the DNN models, deeply embedded in the neurons.
However, a sophisticated DNN contains billions of neurons that interact with each other.
Therefore, it is a challenge to devise an effective algorithm to remove them.
Secondly, since the neural network relies on extracted feature maps to perform desired functions, directly removing them will inevitably degrade the performance of the surrogate model.
As a result, the second challenge for the adversary is to perform an effective attack while maintaining the utility of the surrogate model.
Thirdly, while retraining the model can circumvent ownership verification, it can be a resource-intensive and time-consuming process, particularly when the attacker is unaware of the training data. 
In other words, the attacker should balance the trade-off between efficiency and effectiveness.

In this paper, we investigate the \textit{post-stealing} scenario attacks
    and propose an optimization-based DNN fingerprint removal attack \textsc{RemovalNet}.
We categorize the knowledge of DNNs as general semantic and fingerprint-specific knowledge~\cite{ye2022learning}.
General semantic knowledge is relevant to the main task of the model, 
    which controls the model’s accuracy. 
On the other hand, fingerprint-specific knowledge comes from intrinsic characteristics of the DNN model, 
    which is reflected as a sequence of behavioral patterns on latent representations and decision boundaries.
The key idea of our method is to extract the general semantic knowledge of the victim model 
    and progressively delete its behavioral patterns.
In particular, we propose a min-max bilevel optimization algorithm to 
    remove the fingerprint of the victim model. 
Specifically, in the lower-level problem, 
    we intend to fine-tune a surrogate model that behaves differently from the victim model
    (e.g., layer outputs and activated neurons) but expresses similar semantic meaning. 
In contrast, in the upper-level problem, 
    we seek to learn the victim model's general semantic knowledge to maintain its performance.

We conduct extensive experiments on five benchmark datasets 
    (i.e., CIFAR10, GTSRB, Skin Lesion Diagnosis, CelebA, and ImageNet), 
    including traffic sign recognition, disease diagnosis, face recognition, and large-scale visual recognition scenarios.
The evaluation results demonstrate 
    that the proposed \textsc{RemovalNet} effectively and efficiently removes the DNN fingerprint 
    on the latent representations and decision boundaries.
The major contributions of this paper are summarized as follows:
\begin{itemize}
    \item We perform the first systematic investigation on the DNN fingerprint removal attack, 
        demonstrating that it poses immense threats to model copyright protection.
    \item We propose a min-max bilevel optimization-based DNN fingerprint removal attack. 
    The \textsc{RemovalNet} is primarily designed to 
    remove the victim model's behavioral patterns on latent representations and decision boundaries.
    \item We conduct extensive experiments to evaluate the \textsc{RemovalNet} across six metrics on five benchmark datasets. 
    The evaluation results demonstrate that the \textsc{RemovalNet} can effectively and efficiently remove DNN fingerprints while maintaining high accuracy on the surrogate model.
\end{itemize}

\begin{figure}[!h]
  \centering
  \includegraphics[width=\linewidth]{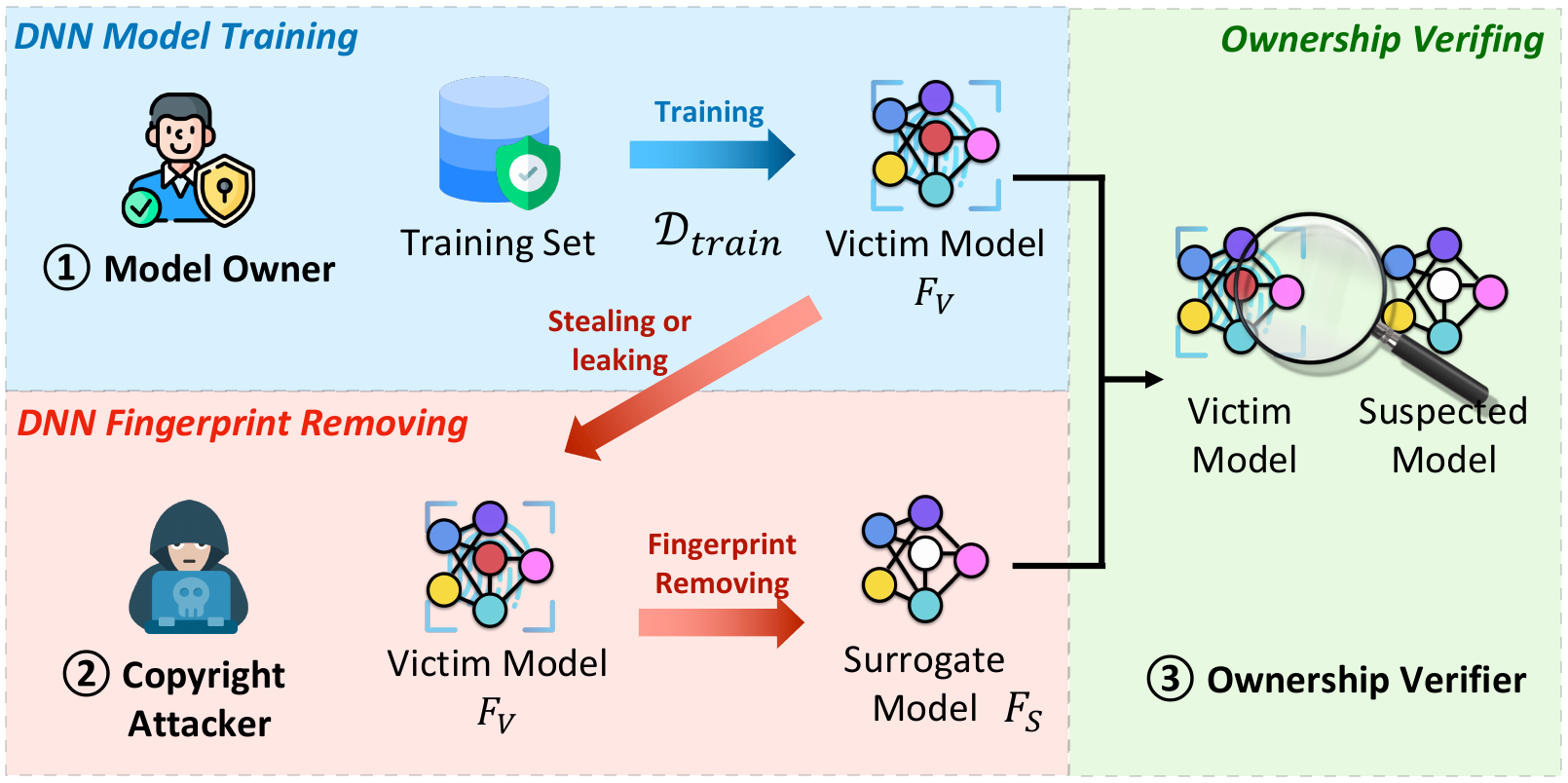}
  \caption{Illustration of three roles in DNN ownership verification, a \textit{model owner}, a \textit{copyright attacker}, and an \textit{ownership verifier}.}
  \label{fig:ownership}
\end{figure}

\section{Preliminaries}
\subsection{Problem Statement}
A Deep Neural Network (DNN) model is a function 
    $F: \mathcal{X} \rightarrow \mathcal{Y}$ 
    parameterized by a set of parameters $\mathbf{w}$, 
    where $\mathcal{X} \in \mathbb{R}^{d}$ is a $d$-dimensional input space 
    and $\mathcal{Y} \in \{1,2,...,K\}$ represents a set of $K$-label.
For the sake of convenience, 
    we divide the original model $F$ into two parts at the $l$-th layers, 
    referring to as $F^{l-}$ and $F^{l+}$, respectively. 
Therefore, the original model is denoted as $F(x) = F^{l+} \circ F^{l-}(x)$,
    where $F^{l-}(x)$ represents the output of layer $l (2 \leq l \leq L)$.

\subsection{Ownership Verification}
In practice, we consider three roles in the problem of ownership verification, 
    i.e., a \textit{model owner}, a \textit{copyright attacker}, and an \textit{ownership verifier} 
    (as illustrated in Figure~\ref{fig:ownership}).
The model owner collects and utilizes training set $\mathcal{D}_{train}$ to build his model ${F}_{V}$ that achieves high accuracy on the test set $\mathcal{D}_{test}$.
The adversary leverages the victim model to derive a surrogate model ${F}_{S}$ aiming to break the ownership verification system.
As a third party, the verifier has white-box access to both the victim model ${F}_{V}$ and the suspected model ${F}_{S}$.

\textbf{DNN fingerprinting.}
DNN fingerprinting relies on intrinsic characteristics of the neural network
  to verify whether the suspected model ${F}_{S}$ is pirated of ${F}_{V}$.
Those characteristics are the traces of training optimization, 
reflected as a sequence of unique model behavioral patterns for specific probing inputs (i.e., fingerprint data).
Without loss of generality, a DNN fingerprinting algorithm consists of two phases:
  \textit{fingerprint extraction} and \textit{copyright verification}.
\begin{itemize}
  \item \textbf{Fingerprint extraction: ${E}({F}_{V}, \mathcal{D}_{train})$.} 
    Given a victim model ${F}_{V}$ and training data $\mathcal{D}_{train}$,
    outputs a number of $N_{fp}$ fingerprints, $\mathcal{D}_{fp}={E}({F}_{V}, \mathcal{D}_{train})$.
  \item \textbf{Copyright verification: ${V}({F}_{V}, {F}_{S}, \mathcal{D}_{fp}, d)$.}
    The copyright verification algorithm exploits lots of distance metrics as evidence to determine the results.
    Given a fingerprint set $\mathcal{D}_{fp}$, a distance metric $d$ returns the average distance between ${F}_{V}$ and ${F}_{S}$.
\end{itemize}

\subsection{Model Distance Metrics}
\label{sec:evaluation_metric}
The advanced DNN fingerprinting methods typically rely on various distance metrics to establish evidence for asserting ownership of a suspected model.
Specifically, those metrics can be categorized into 
    the latent representation-based distance (e.g., DeepJudge~\cite{chen2022copy}, ZEST~\cite{jia2021zest})
    or decision boundary-based similarity (e.g., ModelDiff~\cite{li2021modeldiff}, IPGuard~\cite{cao2021ipguard}.
Additionally, those metrics compass a wide range of cutting-edge techniques, covering white-box and black-box defense strategies. We will describe those metrics in the following paragraph.

\textbf{DeepJudge~\cite{chen2022copy}.}
Chen~\textit{et al.} propose six metrics to measure the distance between two DNN models, 
    which can be used as evidence for ownership verification.
We select two representative metrics to evaluate our fingerprint removal attack, 
    i.e., Layer Output Distance (LOD) and Layer Activation Distance (LAD).
LOD measures the $L_{p}$-norm distance between the two models’ layer outputs:
\begin{equation}
    \resizebox{0.77\hsize}{!}{
    ${LOD}\left({F}_{V}, {F}_{S}, \mathcal{D}_{fp}\right)=
      \frac{1}{N_{fp}} \sum_{i = 1}^{N_{fp}}
        \left\|{F}_{V}^l({x}_{i})-{F}_{S}^l({x}_{i})\right\|_p $.
    }
  \end{equation}
LAD measures the average distance of activation neurons for layer $l$:
\begin{equation}
    \resizebox{0.9\hsize}{!}{
    ${LAD} \left({F}_{V}, {F}_{S}, \mathcal{D}_{fp}\right)=
      \frac{1}{N_{l} \times N_{fp}} 
        \sum_{j = 1}^{N_{l}}
          \sum_{i = 1}^{N_{fp}} 
            |{\phi_{T}^{l, j}}({x}_{i}) - {\phi_{S}^{l, j}}({x}_{i}) |$,
    }
\end{equation}
where $N_{l}$ denotes the number of neurons for $l$-th layer's feature maps,
$\phi_{T}^{l, i}$ returns 1 if the $i$-th neuron for the output of $l$-th layer greater than a certain threshold.
Since the model distance computation is based on the output of intermediate layers, 
    the two suspected models should have the same model architecture.

\textbf{ZEST~\cite{jia2021zest}.}
Jia~\textit{et al.} leverage the LIME~\cite{ribeiro2016should} technique to approximate the global behavior of two models.
ZEST first approximates the models on reference data with linear models trained by LIME.
Subsequently, ZEST computes the L2 norm or cosine similarity 
    on the trained models to quantitatively measure the distance between two suspected models.

\textbf{ModelDiff~\cite{li2021modeldiff}.} 
Li~\textit{et al.} propose Decision Distance Vector (DDV) to approximate the behavioral patterns of a model:
\begin{equation}
    DDV_{F}(\mathcal{D}_{fp}) = \{||F(x_{i}) - F(x_{j})||_{2} | 0 \le i <  j \leq N_{fp} \}.
\end{equation}
Afterward, the knowledge similarity between two models 
    is measured with the cosine similarity between their DDVs:
\begin{small}
\begin{gather}
    Sim({F}_{V}, {F}_{S}) = CosineSimilarity(DDV_{{F}_{V}}, DDV_{{F}_{S}}).
\end{gather}
\end{small}

\textbf{IPGuard~\cite{cao2021ipguard}.}
IPGuard explores data points near the decision boundaries to fingerprint the boundary property of the victim model.
In this context, Marching Rate (MR) is proposed as a metric to measure the degree of similarity between two models on the decision boundary:
\begin{equation}
    \resizebox{0.85\hsize}{!}{
    $ MR = \frac{1}{N_{fp}} \sum_{i=1}^{N_{fp}} 
      \mathbbm{1} \left[\arg \max {F}_{V}(x_{i}) 
        = \arg \max {F}_{S}(x_{i})\right]$.
    }
\end{equation}
It should be noted that both DeepJudge and ZEST return the distance between two models, 
    while the ModelDiff and IPGuard return the similarity score of the two models.

\begin{figure*}[!h]
  \centering
  \includegraphics[width=0.72\linewidth]{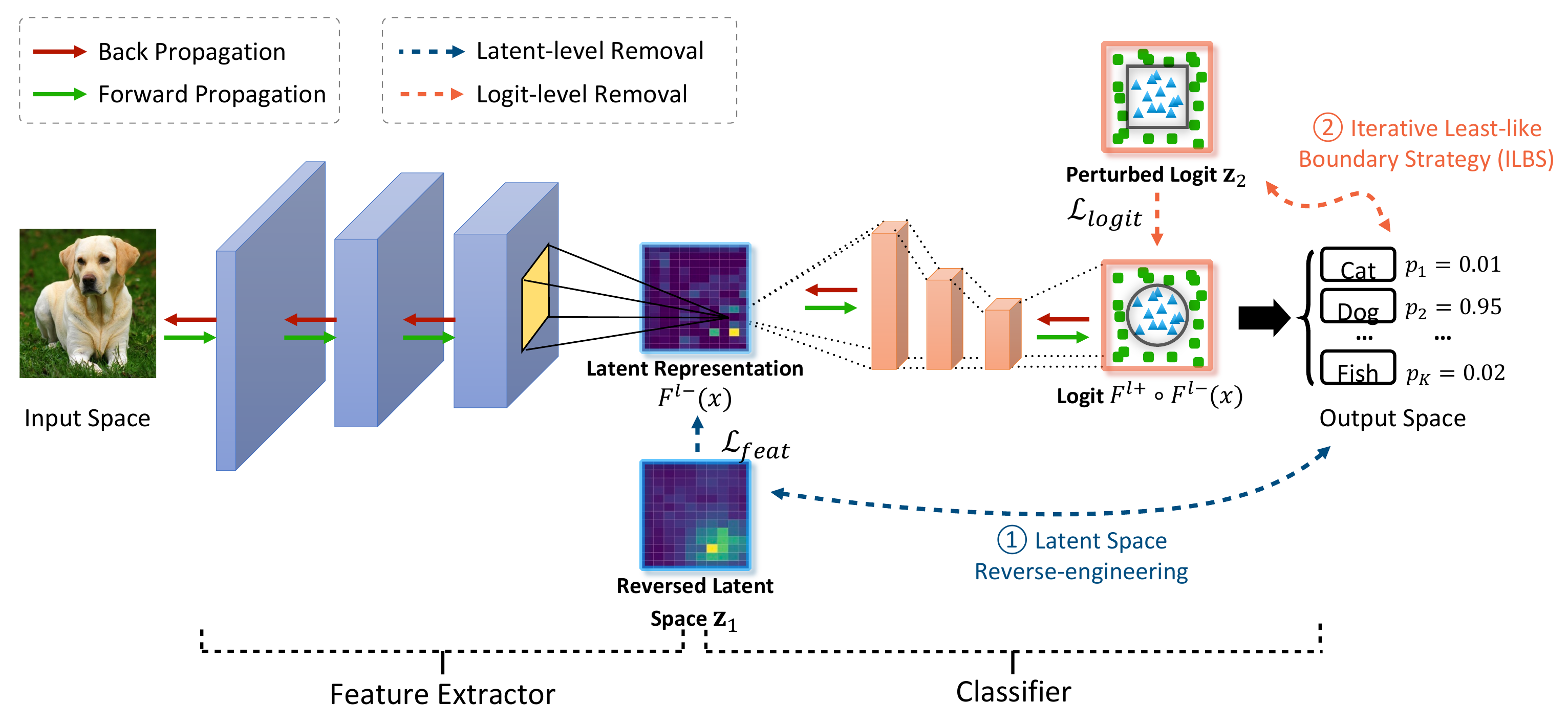}
  \caption{Overview of \textsc{RemovalNet} against DNN ownership verification.
  The removal process is conducted on the latent-level and logit-level to alter the behavior patterns in the latent representation and decision boundary, respectively.}
  \label{fig:pipeline}
\end{figure*}

\subsection{Threat Model}
\label{sec:threat_model}
In this section, we formulate the threat model to characterize the adversary’s goals and capabilities. We will first define DNN fingerprint removal attacks. \par

\partitle{DNN fingerprint removal attacks}
Given a victim model ${F}_{V}$ and a substitute set $\mathcal{D}_{sub}$, 
    the adversary intends to derive a surrogate model ${F}_{S}$ from $F_{V}$.
The DNN fingerprint removal attacks are designed to purge fingerprint-specific knowledge of the victim model. 
After the removal procedure, the created surrogate model is expected to behave differently from the victim model 
    (e.g., output distance or distribution of activated neurons or decision boundaries).
The formal definition of DNN fingerprint removal attacks is given below:
\begin{definition}[DNN fingerprint removal]
    \label{def:fp}
    Given a victim model ${F}_{V}$, a fingerprint set $\mathcal{D}_{fp}$,
    a copyright verification method $V$,
    the surrogate model ${F}_{S}$ is $(\mathcal{D}_{fp}, d, \tau)$\textit{-fingerprint removal} if
    \begin{equation}
        \mathbb{E}_{x \sim \mathcal{D}_{fp}} [V\left({F}_{V}, {F}_{S}, x, d\right)] > \tau,
    \end{equation}
    where $d$ denotes the distance metric, $\tau$ is a threshold.
    Noted that the adversary can't access $D_{fp}$ but uses a substitute set $D_{sub}$ to conduct the removal procedure.
\end{definition}

\partitle{Attack motivations}
We first list potential motivations for conducting the DNN fingerprint removal attack.
\textit{(1) Compromising system reliability.}
The adversary intends to derive a functionality-preserved surrogate model ${F}_{S}$ from the victim model ${F}_{V}$, 
    which obfuscates the defense algorithm and breaks the reliability of the ownership verification system.
Besides, the removal attack should remarkably reduce the training resources 
    (e.g., computational resources and data for training) compared with model retraining.
Therefore, the adversary’s goal falls into three categories, 
  i.e., \textit{fidelity}, \textit{effectiveness}, and \textit{efficiency}.
\begin{itemize}
    \item \textbf{Fidelity:}
    The surrogate model should maintain the performance of the victim model on the $\mathcal{D}_{test}$. 
    We adopt accuracy and accuracy drops to measure the fidelity of the surrogate model.
    Therefore, the goal of the adversary in fidelity can be summarized as follows:
    \begin{equation}
      \max_{{F}_{S}} {P}_{(x, y) \in \mathcal{D}_{test}} 
        \mathbbm{1}\left[\arg\max{{F}_{S}(x)} = y\right].
    \end{equation}
    \item \textbf{Effectiveness:} 
    The effectiveness metric describes the capability of DNN fingerprint removal attacks to evade ownership verification techniques.
    As a result, the goal of the adversary in effectiveness can be summarized as follows:
    \begin{equation}
      \max_{{F}_{S}} {P}_{x \in \mathcal{D}_{fp}} 
        [{V}\left({F}_{V}, {F}_{S}, x, d\right)].
    \end{equation}
    \item \textbf{Efficiency:}
    In general, retraining a model can circumvent ownership verification completely. 
    While retraining has proven successful, there is a trade-off between its effectiveness and efficiency.
    Generally, DNN fingerprint removal attacks cost significantly less than model retraining.
\end{itemize}

\textit{(2) Understanding the capability of attack.}
Except for the aforementioned malicious goal, we consider a more open-world scenario to evaluate the capability of DNN fingerprint attacks. 
Specifically, the adversary attempts to determine the minimum resources required to bypass a DNN ownership verification system, which can also help evaluate the system's overall robustness. 
Additionally, we suggest that DNN fingerprint removal attacks can be leveraged as a novel tool for evaluating the performance of ownership verification systems. 
For example, a confidential ownership verification system can utilize our algorithm to comprehensively evaluate its system before deployment, guaranteeing its effectiveness and security.

\partitle{Attacker's capabilities}
Motivated by recent works in DNN watermark removal attacks~\cite{lukas2022sok,guo2020fine,chen2021refit}, 
    we make the assumption that the adversary has a limited substitute set $\mathcal{D}_{sub}$ for fine-tuning.
The substitute set assumption is realistic since many organizations or research communities have released plenty of benchmark datasets, part of which can be utilized as surrogate data.
Specifically, we consider two scenarios of the adversary:
\begin{itemize}
    \item \textbf{Limited Training Data ($\mathcal{S}_{LTD}$).} 
        The adversary only has access to a small part of training data (e.g., 10\% of training data).
    \item \textbf{Limited Surrogate Data ($\mathcal{S}_{LSD}$).}
        The adversary may acquire natural samples that share a similar distribution to the training set.
\end{itemize}

\partitle{Defender’s capabilities}
The defender has full access to both the victim model ${F}_{V}$, the suspected model ${F}_{S}$,
  and the fingerprint set $\mathcal{D}_{fp}$.
The defender has unlimited computation resources to 
    extract fingerprints and verify the ownership of the suspected models.

\section{\textsc{RemovalNet}}
\label{sec:method}
In this section, we present the details of \textsc{RemovalNet}, including substitute set selection, latent-level removal, and logit-level removal.

\partitle{Overall workflow}
Our approach divides the neural network into two modules: 
    a feature extractor that produces latent representations and a classifier that returns the logit vector. 
We propose removing DNN fingerprints at the latent and logit levels, as depicted in Figure~\ref{fig:pipeline}. 
The latent-level removal is intended to eliminate the behavior patterns in the representation layer, while the logit-level removal alters the decision boundaries’ behavior patterns.

Firstly, we reverse the latent layer from the victim model and inject perturbation into it:
\begin{equation}
    \{\mathbf{z}_{1}, \mathbf{z}_{2}\} = \{{F}_{V}^{l-}(x) + \delta_{1}, {F}_{V}^{l+} \circ {F}_{V}^{l-}(x) + \delta_{2}\}, 
\end{equation}
where $\delta_{1}$ and $\delta_{2}$ perturb on the latent representation layer $l$ and the penultimate layer, respectively.
Secondly, we minimize the distance between $\{\mathbf{z}_{1}, \mathbf{z}_{2}\}$ and the surrogate model’s corresponding outputs.
Subsequently, the behavioral patterns of the victim model will progressively be removed from the surrogate model.

The removal procedure can be formulated as a min-max bilevel optimization problem.
In our method, the lower-level problem is to find a layer output 
    that behaves differently from the victim model but has a similar semantic meaning.
This process is conducted by maximizing the distance of latent representation between the target and surrogate models.
In contrast, the upper-level problem involves distilling the reversed layer output 
    while minimizing the loss of the main task to prevent catastrophic forgetting.
Specifically, given a loss function $\mathcal{L}$,
    the optimal surrogate model ${F}_{S}$ can be optimized by empirical risk:
\begin{align}
\label{eq:minmax_objective}
   \mathbf{\mathbf{w}} =& \min_{\mathbf{w}} \max_{\mathbf{z}}
        \mathcal{L}({\mathbf{w}, \{\mathbf{z}^{*}\}; \mathcal{D}_{sub}}), \\ 
    s.t. & \arg\max {F}_{S}^{l+}(\mathbf{z}^{*}) = \arg\max {F}_{V}^{l+}(\mathbf{z}^{*}), \notag
\end{align}
where $\mathbf{w}$ is the parameters of surrogate model ${F}_{S}$, 
$\mathbf{z}$ denotes the output of selected layer $l$.
Figure~\ref{fig:pipeline} schematically illustrates the framework of \textsc{RemovalNet}.

\subsection{Substitute Set Selection}
As described in Definition~\ref{def:fp}, the DNN ownership is validated on the fingerprint set $\mathcal{D}_{fp}$.
However, the DNN fingerprint set is generated and safeguarded by the verifier.
Therefore, the first step toward the DNN fingerprint removal attacks is to select the substitute set $\mathcal{D}_{sub}$.

The selection of the substitute set should adhere to two key principles: 
    (1) diversity in classes and (2) similarity in data distribution. 
The ownership verification algorithms are usually designed by 
    measuring the distances between the target and surrogate models over different classes. 
We select substitute data with diverse classes to enhance the removal attacks’ performance. 
Additionally, we choose natural samples that reflect a similar distribution to the training set to avoid the risk of catastrophic forgetting. 
Note that we only use a small number of substitute data, which can be collected from the public benchmark datasets.
We also discuss the confidential scenario in Section~\ref{sec:substitute_ratio}, where collecting the substitute data is challenging.

\subsection{Latent-level Removal}
\begin{figure}[!th]
  \centering
  \includegraphics[width=0.84\linewidth]{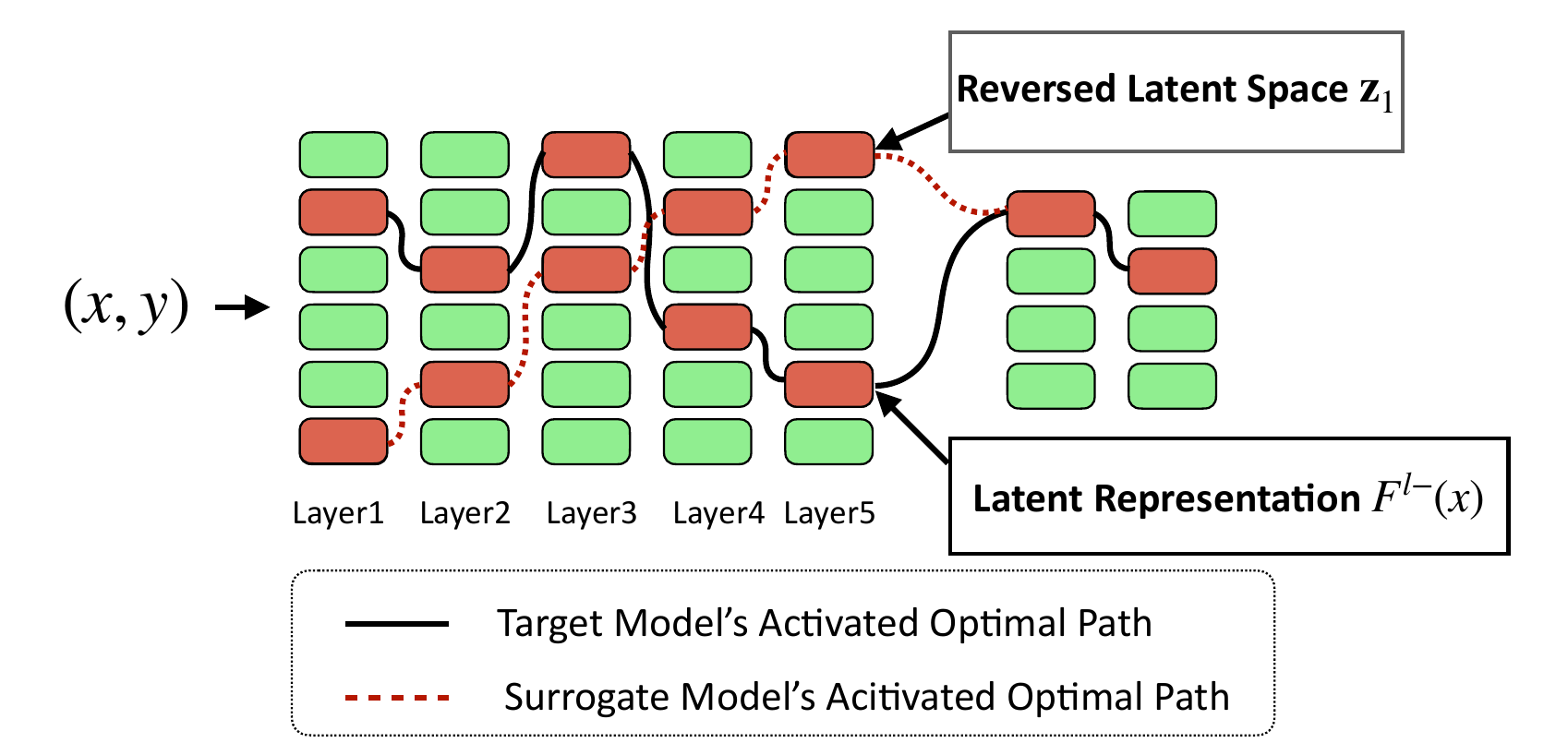}
  \caption{
  DNN fingerprint removal attack on latent-level. 
  The reversed latent space follows a logic similar to the victim model but triggers an alternative path.}
  \label{fig:sec4_feature_space}
\end{figure}
The artificial neural network expresses logic using a sequence of activated neurons 
    that output values higher than the threshold~\cite{pei2017deepxplore}.
Since the activation status of neurons is highly related to the learning objective and learning progress, 
    it can be exploited to measure the model distance. 
Therefore, the activation status of neurons plays an important role in DNN fingerprinting.
Then, we would ask whether we can find an optimal path 
    that activates a set of neurons different from the victim model.
Recent works on mitigating catastrophic forgetting have investigated 
    the possibility of finding another optimal path for new tasks~\cite{mallya2018packnet,rajasegaran2019random}.
Those works exploit the redundant weights of neural networks to activate a different set of neurons (as illustrated in Figure~\ref{fig:sec4_feature_space}).

\begin{algorithm}
\small
\setstretch{1.2}
\caption{Latent space reverse-engineering}
\label{alg:feats_reverse}
\KwIn{victim model ${F}_{V}$, surrogate model $F_{S}$,
    selected layer $l$, 
    ${F}_{V}$’s first $l$ layer output ${F}_{V}^{l-}$,
    ${F}_{V}$’s last $l$ layer output ${F}_{V}^{l+}$,
    substitute data $(x, y) \in \mathcal{D}_{sub}$, 
    learning rate $\eta$.}
\KwOut{Feature maps hyperplane $\mathbf{z}_{1}$}
$\mathbf{z}_{1}^{\prime} = {F}_{V}^{l-}(x)$ \\
$\hat{y} = \arg\max {F}_{V}(x)$ \\
\For{$i \gets 1$ \textbf{to} 10}{
    $\mathbf{z}_{1} = \textit{Feature\_Shuffling}(\mathbf{z}_{1}, ratio=0.1)$ \\
    $\mathcal{L}_{re} = 
        \mathcal{L} _{CE}({F}_{S}^{l+}(\mathbf{z}_{1}), \hat{y})
        -\log {||\mathbf{z}_{1}-\mathbf{z}_{1}^{\prime}||_{2}} $ \\
    $\mathbf{z}_{1} = \mathbf{z}_{1} 
        - \eta \cdot \frac{\partial \mathcal{L}_{re}}{\partial \mathbf{z}_{1}}$ \\
}
\Return{$\mathbf{z}_{1}$}
\end{algorithm}
Based on the abovementioned insight, we propose the latent-level fingerprint removal attack.
The first step toward latent-level fingerprint removal is to reverse a detached latent space $\mathbf{z}_{1}$.
Similar to reverse engineering in the input space, 
    we reverse the latent space hyperplane $\mathbf{z}_{1}$ with two constraints: 
    (1) adopt cross-entropy to make $\mathbf{z}_{1}$ have similar semantic logic as $F_{V}^{l-}(x)$,
    (2) maximize the distance between $\mathbf{z}_{1}$ and ${F}_{V}^{l-}(x)$
    (as illustrated in line 5 of Algorithm~\ref{alg:feats_reverse}).
To enhance diversity during optimization, we employ feature shuffling on the reversed latent representation inspired by recent data transformation works (line 4 of Algorithm~\ref{alg:feats_reverse}).

In contrast, in the upper-level problem, we optimize the following loss function:
\begin{align}
    \mathcal{L}_{feat} & = ||{F}_{S}^{l-}(x) - \mathbf{z}_{1}||_{2}.
\label{eq:upper_loss_feat}
\end{align}
Finally, we project the reversed latent space $\mathbf{z}_{1}$ on the upper-level problem using the following objective function:
\begin{align}
        \mathbf{w} = &\min_{\mathbf{w}} \max_{\mathbf{z}_{1}^{*}} \mathcal{L}_{feat} \\
    s.t. & \arg\max {F}_{S}^{l+}(\mathbf{z}_{1}^{*}) = \arg\max {F}_{V}(x). \notag
\label{eq:upper_feat}
\end{align}

\begin{algorithm}[!ht]
\small
\setstretch{1.2}
\caption{Iterative least-like boundary strategy}
\label{alg:boundary_poisoning}
\KwIn{victim model ${F}_{V}$,
    a batch of substitute data $\mathcal{B} \in \mathcal{D}_{sub}$.}
\KwOut{Perturbed logit ${\mathbf{z}_{\mathcal{B}}}$}
${\mathbf{z}_{\mathcal{B}}} = \{\}$, ${idx_{\mathcal{B}}} = \{\}$ \\
// Find the indexes of the greatest distance samples\\
\For{$i \gets 1$ \textbf{to} $|\mathcal{B}|$}{
    $(x_{i}, y_{i}) = \mathcal{B}[i]$ \\
    ${d_{max}} = 0, idx_{max} = i$ \\
    \For{$j \gets 1 $ \textbf{to} $|\mathcal{B}|$}{
        $(x_{j}, y_{j}) = \mathcal{B}[j]$ \\
        \If{$d_{max} < ||{F}_{V}(x_{i}) - {F}_{V}(x_{j})||_{2}$}{
            $d_{max} = ||{F}_{V}(x_{i}) - {F}_{V}(x_{j})||_{2}$ \\
            $idx_{max} = j$
        }
    }
    ${idx_{\mathcal{B}}}[i] = idx_{max} $ \\
}
// Interpolate the target logit to maximize the decision boundary deviation \\
\For{$i \gets 1$ \textbf{to} $|\mathcal{B}|$}{
    $j = {idx_{\mathcal{B}}}[i]$ \\
    $(x_{i}, y_{i}) = \mathcal{B}[i]$ \\
    // Line search to find the optimal $\lambda$ \\
    \For{$\lambda = \{0.1,0.2,...,1.0\}$}{
        $\mathbf{z}_{2} = \lambda \cdot {F}_{V}(x_{i})
            + (1-\lambda) \cdot {F}_{V}(x_{j})$ \\
        \If{$\arg\max \mathbf{z}_{2} = \arg\max {F}_{V}(x)$}{
           ${\mathbf{z}_{\mathcal{B}}}[i] = {\mathbf{z}_{2}}$ \\
        }
    }
}
\Return{${\mathbf{z}_{\mathcal{B}}}$}
\end{algorithm}
\begin{table*}[!ht]
    \centering
    \caption{Performance of \textsc{RemovalNet} against DNN ownership verification.
    Large is better for DeepJudge and ZEST, while small is better for ModelDiff and IPGuard.
    ``DJ-LOD'' denotes the LOD metric of DeepJudge.}
    \resizebox{\linewidth}{!}{
    \begin{tabular}{c|c|cc|c|c|c|c|c|c|c}
    \specialrule{1pt}{0pt}{0pt}
    \multirow{2}{*}{\textbf{Dataset}}
    & \multirow{2}{*}{\textbf{Attack Model}}
    & \multicolumn{2}{c|}{\textbf{Fidelity (\%)}}
    & \multicolumn{4}{c|}{\textbf{Distance} ($\uparrow$ \textit{better})} 
    & \multicolumn{2}{c}{\textbf{Similarity} ($\downarrow$ \textit{better})} & \multirow{2}{*}{\textbf{Ownership?}}\\
    && ${F}_{S}$ & Drop$\downarrow$ &
    DJ-LOD & 
    DJ-LAD & 
    ZEST-L2 &
    ZEST-Cosine &
    ModelDiff-DDV & 
    IPGuard-MR & \\
    \specialrule{1pt}{0pt}{0pt}
    \multirow{9}{*}{\tabincell{c}{\textbf{CIFAR10} \\ \textbf{(Acc: 90.58\%)}}}
    &FT(0.5)    &88.21±0.31&2.05    
        & 0.01±0.02 & 0.04±0.03
        & 0.76±0.14 & 0.01±0.01    
        & 0.998±0.002 & 0.885±0.020 &\textbf{Yes} (10/10)\\
    &FT(0.8)    &87.22±0.54&3.03    
        & 0.02±0.02   & 0.04±0.04
        & 1.03±0.15   & 0.01±0.01    
        & 0.997±0.003 & 0.875±0.031 &\textbf{Yes} (10/10)\\
    &WP(0.5)    &87.25±0.63&4.03    
        & 0.01±0.02 & 0.04±0.02
        & 2.84±0.47 & 0.01±0.01
        & 0.993±0.001 & 0.875±0.073 &\textbf{Yes} (10/10)\\
    &WP(0.8)    &85.27±0.63&5.21    
        & 0.02±0.03  & 0.05±0.02
        & 14.04±0.49 & 0.02±0.01
        & 0.943±0.027 & 0.844±0.031 &\textbf{Yes} (10/10)\\
    &Distill    &86.37±0.56&4.05    
        & 0.04±0.03   & 0.03±0.02
        & 8.28±4.96   & 0.02±0.01
        & 0.930±0.022 & 0.885±0.166 &\textbf{Yes} (10/10)\\
    &\underline{Ours($\mathcal{S}_{LTD}$)}
        & \underline{87.17±1.75} & \underline{2.89}    
        & \underline{\textbf{12.62±3.60}}  & \underline{5.11±0.69}
        &\underline{\textbf{115.86±2.71}} &\underline{0.51±0.03}
        & \underline{0.321±0.140} & \underline{0.061±0.043} &\textbf{No} (0/10)\\
    &\underline{Ours($\mathcal{S}_{LSD}$)}
        &\underline{86.81±2.24} &\underline{3.23}    
        &  \underline{8.42±0.41}  &  \underline{\textbf{5.51±0.69}}
        &  \underline{101.86±4.42} &  \underline{\textbf{0.52±0.12}}    
        & \underline{\textbf{0.211±0.162}} & \underline{\textbf{0.002±0.022}} &\textbf{No} (0/10)\\
    &Negative   &86.69±1.08 &3.75    
        & 10.72±2.00 & 4.99±0.90
        & 90.76±3.09 & 0.40±0.02
        & 0.325±0.231&  0.010±0.021 &\textbf{No} (0/10)\\
    \cline{2-11}
    & $\tau$ & - & - & 6.49 & 2.95 & 75.12 & 0.319 & 0.450 & 0.250 & - \\
    \hline

    \multirow{9}{*}{\tabincell{c}{\textbf{GTSRB} \\ \textbf{(Acc: 92.26\%)}}}
    &FT(0.5)    & 94.03±0.25 & 0.25    
        & 3.23±0.26 & 0.28±0.01
        & 8.78±0.87 & 0.01±0.01   
        & 0.959±0.029 & 0.857±0.071 &\textbf{Yes} (10/10)\\
    &FT(0.8)    &93.16±0.30&1.03    
        & 3.40±0.26 & 0.28±0.01
        & 10.60±1.08 & 0.01±0.00
        & 0.958±0.029 & 0.857±0.143 &\textbf{Yes} (10/10)\\
    &WP(0.5)    &92.94±0.43&1.25    
        & 3.69±0.24 & 0.38±0.01
        & 16.19±0.77 & 0.03±0.00
        & 0.935±0.031 & 0.857±0.143 &\textbf{Yes} (10/10)\\
    &WP(0.8)    &92.01±0.82&2.16    
        & 3.97±0.31 & 0.22±0.01
        & 19.34±2.88 & 0.13±0.01
        & 0.576±0.112 & 0.429±0.286 &\textbf{Yes} (10/10)\\
    &Distill    &91.78±1.64&2.23   
        & 3.44±0.27 & 0.27±0.02
        & 21.74±3.11 & 0.07±0.02
        & 0.489±0.186 & 0.714±0.071  &\textbf{Yes} (10/10)\\
    
    &\underline{Ours($\mathcal{S}_{LTD}$)}    
        &\underline{91.88±1.67}&\underline{2.01}
        &\underline{\textbf{11.20±0.05}}&\underline{19.92±0.30}
        &\underline{137.01±0.12}&\underline{0.63±0.04}
        &\underline{\textbf{0.115±0.146}}&\underline{0.251±0.143} &\textbf{No} (0/10)\\
    
    &\underline{Ours($\mathcal{S}_{LSD}$)}    
        &\underline{92.02±0.75}&\underline{2.15}
        &\underline{11.19±0.04}&\underline{\textbf{19.77±0.21}}
        &\underline{135.12±0.05}&\underline{\textbf{0.65±0.01}}
        &\underline{0.118±0.114}&\underline{0.286±0.214} &\textbf{No} (0/10)\\
    &Negative   &92.12±1.88&2.18 
        &9.05±1.18&15.99±1.10
        &\textbf{138.21±9.19}&0.59±0.11
        &0.132±0.243&\textbf{0.242±0.021}  &\textbf{No} (0/10)\\
    \cline{2-11}
    & $\tau$ & - & - & 6.02 & 13.26 & 92.13 & 0.40 & 0.293 & 0.472 & - \\
    \hline
    
    \multirow{8}{*}{\tabincell{c}{\textbf{Skin Lesion} \\ \textbf{(Acc: 98.81\%)}}}
    &FT(0.5)    &96.90±0.14&2.06    &0.05±0.01&0.03±0.01&1.84±0.10&0.02±0.01    &0.982±0.001&0.902±0.009 &\textbf{Yes} (10/10)\\
    &FT(0.8)    &95.72±0.24&3.16    &0.08±0.02&0.04±0.03&2.34±0.14&0.11±0.03    &0.958±0.004&0.915±0.012 &\textbf{Yes} (10/10)\\
    &WP(0.5)    &96.80±0.15&2.15    &0.07±0.06&0.06±0.01&5.10±0.08&0.04±0.01    &0.995±0.003&0.853±0.011 &\textbf{Yes} (10/10)\\
    &WP(0.8)    &95.64±0.13&4.31     &0.10±0.07&0.07±0.03&12.09±2.24&0.02±0.01    &0.957±0.014&0.856±0.062 &\textbf{Yes} (10/10)\\
    &Distill    &95.01±0.41&4.06    &0.08±0.02&0.06±0.02&10.35±2.02&0.14±0.02    &0.951±0.005&0.849±0.048 &\textbf{Yes} (10/10)\\
    &\underline{Ours($\mathcal{S}_{LTD}$)} 
        &\underline{97.29±0.42} &\underline{1.21}
        &\underline{\textbf{9.58±0.26}}  &\underline{3.99±1.05}
        &\underline{67.94±3.51} &\underline{0.48±0.12}
        &\underline{0.159±0.125}&\underline{0.164±0.044} &\textbf{No} (0/10)\\
    &\underline{Ours($\mathcal{S}_{LSD}$)}  
        &\underline{94.35±1.38}&\underline{3.34}
        &\underline{8.14±0.84} &\underline{\textbf{3.32±0.63}}
        &\underline{75.32±5.22}     &\underline{\textbf{0.54±0.12}}
        &\underline{\textbf{0.152±0.156}}   &\underline{0.161±0.036} &\textbf{No} (0/10)\\
    &Negative   &98.58±0.43&0.25
        &9.57±1.84&4.09±0.63
        &\textbf{86.27±9.22}&0.52±0.07
        &0.163±0.039&\textbf{0.035±0.031} &\textbf{No} (0/10)\\
    \cline{2-11}
    & $\tau$ & - & - & 6.37 & 2.72 & 57.45 & 0.35 & 0.361 & 0.405 & - \\
    \specialrule{1pt}{0pt}{0pt}
    \end{tabular}
}
\label{tab:exp11_main_result1}
\end{table*}

\subsection{Logit-level Removal}
Generally, the natural sample is situated close to the center of its respective class, maintaining a significant distance from the decision boundary.
In contrast, the extracted fingerprint data, often from adversarial inputs, tend to lie alongside the decision boundary.
Therefore, it is possible to obfuscate the DNN fingerprinting verification by introducing deceptive non-robustness features.
While these non-robustness features compromise the surrogate model's robustness, they can effectively evade DNN fingerprinting verification.
Specifically, we introduce a slight perturbation to the logit vector, resulting in an altered decision boundary.

The critical challenge in logit-level removal is how to find the perturbated $\mathbf{z}_{2}$.
To tackle this challenge, we propose an iterative algorithm called the \textit{iterative least-like boundary strategy (ILBS)}.
Specifically, the proposed iterative strategy involves identifying pairs in a mini-batch with the greatest distance between logit. 
This is followed by adopting a linear interpolation logit vector to meet the imposed constraints.
We rewrite the perturbed logit as 
    ${F}_{V}(x_{i}) + \delta_{2} = \lambda {F}_{V}(x_{i}) + (1-\lambda) {F}_{V}(x_{j})$,
    where $x_{j}$ is the sample that has the greatest logit distance with $x_{i}$.
Finally, we employ a line search strategy to find the best $\lambda$ for the constraints.
We illustrate the detail of \textit{ILBS} in Algorithm~\ref{alg:boundary_poisoning}.
In lines 3-14 of Algorithm~\ref{alg:boundary_poisoning}, 
    we search for the pairs with the most significant logit distance in a mini-batch.
In lines 16-26 of Algorithm~\ref{alg:boundary_poisoning},
    we project the logit interpolation to maximize the decision boundary deviation.
In lines 20-25 of Algorithm~\ref{alg:boundary_poisoning}, 
    we adopt the linear search to iteratively find the best $\lambda$ 
    that satisfies the constraint of $\mathbf{z}_{2}$.

Additionally, we adopt a cross-entropy loss on the logit-level removal to avoid catastrophic forgetting.
Finally, the loss function can be formulated as:
\begin{equation}
    \mathcal{L}_{logit} = 
        \alpha \cdot \mathcal{L}_{CE}({F}_{V}(x), \hat{y}) + 
        (1-\alpha) \cdot \mathcal{L}_{KL}({F}_{V}(x),\mathbf{z}_{2}), 
\end{equation}
where $\hat{y}=\arg\max {F}_{V}(x)$ is the predicted label.
Then, we project the reversed logit $\mathbf{z}_{2}$ on the upper-level problem using the following objective function:
\begin{align}
    \mathbf{w} = &\min_{\mathbf{w}} \max_{\mathbf{z}_{2}^{*}} \mathcal{L}_{logit} \\
    s.t. & \arg\max \mathbf{z}_{2}^{*} = \arg\max {F}_{V}(x). \notag
\label{eq:logit_removal}
\end{align}

Finally, the loss function in Equation~\ref{eq:minmax_objective} can be rewritten as follows:
\begin{equation}
    \mathcal{L} = \alpha \mathcal{L}_{CE} + (1-\alpha) \mathcal{L}_{KL} + \beta \mathcal{L}_{feat} .
\end{equation}
The latent-level and logit-level optimization are performed simultaneously during the min-max process.

\begin{table*}[!t]
    \centering
    \caption{Performance of \textsc{RemovalNet} against DNN ownership verification.
    Large is better for DeepJudge and ZEST, while small is better for ModelDiff and IPGuard.
    ``DJ-LOD'' denotes the LOD metric of DeepJudge.
    ``CelebA+20'' and ``CelebA+31'' indicate the gender and smiling attributes of CelebA, respectively.
    ZEST fails to work for large model like ViT\_B/32.}
    \resizebox{\linewidth}{!}{
    \begin{tabular}{c|c|cc|c|c|c|c|c|c|c}
    \specialrule{1pt}{0pt}{0pt}
    \multirow{2}{*}{\textbf{Dataset}}
    & \multirow{2}{*}{\textbf{Attack Model}}
    & \multicolumn{2}{c|}{\textbf{Fidelity (\%)}}
    & \multicolumn{4}{c|}{\textbf{Distance} ($\uparrow$ \textit{better})} 
    & \multicolumn{2}{c|}{\textbf{Similarity} ($\downarrow$ \textit{better})} & \multirow{2}{*}{\textbf{Ownership?}}\\
    && ${F}_{S}$ & Drop$\downarrow$ &
    DJ-LOD & 
    DJ-LAD & 
    ZEST-L2 &
    ZEST-Cosine &
    ModelDiff-DDV & 
    IPGuard-MR & \\
    \specialrule{1pt}{0pt}{0pt}
    \multirow{8}{*}{\tabincell{c}{\textbf{CelebA+20} \\ \textbf{(Acc: 97.68\%)}}}
    &FT(0.5)    &95.31±0.04&2.02    &0.02±0.01&0.01±0.01&1.55±0.53&0.01±0.01    &0.991±0.002&0.976±0.011    &\textbf{Yes} (10/10)\\
    &FT(0.8)    &94.23±0.07&3.12    &0.03±0.02&0.02±0.01&3.65±1.23&0.03±0.01    &0.994±0.013&0.967±0.017    &\textbf{Yes} (10/10)\\
    &WP(0.5)    &94.42±0.09&3.06    &0.03±0.01&0.06±0.03&4.62±1.03&0.03±0.02    &0.990±0.011&0.963±0.031    &\textbf{Yes} (10/10)\\
    &WP(0.8)    &93.04±0.07&4.20    &0.04±0.03&0.07±0.02&6.84±0.47&0.06±0.01    &0.985±0.013&0.952±0.023    &\textbf{Yes} (10/10)\\
    &Distill    &93.69±0.20&4.02    &0.08±0.07&0.23±0.10&7.20±1.23&0.05±0.02    &0.960±0.025&0.750±0.091    &\textbf{Yes} (10/10)\\
    &\underline{Ours($\mathcal{S}_{LTD}$)} 
        &\underline{94.18±5.91}&\underline{1.42}
        &\underline{\textbf{9.82±0.46}}&\underline{6.30±1.16}&\underline{32.20±0.64}&\underline{0.18±0.02}
        &\underline{0.516±0.090}&\underline{0.299±0.073}    &\textbf{No} (0/10)\\
    &\underline{Ours($\mathcal{S}_{LSD}$)}  
        &\underline{94.30±3.21}&\underline{3.59}
        &\underline{5.13±0.27}&\underline{\textbf{7.51±1.89}}&\underline{38.79±1.02}&\underline{\textbf{0.35±0.03}}
        &\underline{\textbf{0.378±0.318}}&\underline{0.247±0.172}   &\textbf{No} (0/10)\\
    &Negative   &96.71±0.47&0.94    &7.69±2.03&5.92±1.07&\textbf{46.36±4.02}&0.26±0.03    &0.418±0.294&\textbf{0.039±0.015} &\textbf{No} (0/10) \\
    \cline{2-11}
    & $\tau$ & - & - & 5.14 & 3.94 & 30.87 & 0.18 & 0.593 & 0.410 & - \\
    \hline
    \multirow{8}{*}{\tabincell{c}{\textbf{CelebA+31} \\ \textbf{(Acc: 92.50\%)}}}
    &FT(0.5)    &91.24±0.12&1.09    &0.06±0.04&0.07±0.03&1.27±0.21&0.01±0.01    &0.991±0.012&0.978±0.012    &\textbf{Yes} (10/10)\\
    &FT(0.8)    &90.17±0.11&2.07    &0.07±0.05&0.08±0.03&2.32±0.23&0.02±0.01    &0.975±0.008&0.944±0.020    &\textbf{Yes} (10/10)\\
    &WP(0.5)    &90.65±0.17&1.85    &0.09±0.06&0.09±0.04&2.64±0.12&0.02±0.01    &0.989±0.013&0.962±0.021    &\textbf{Yes} (10/10)\\
    &WP(0.8)    &90.21±0.23&2.28    &0.10±0.08&0.10±0.01&6.43±0.12&0.03±0.09    &0.981±0.009&0.951±0.008     &\textbf{Yes} (10/10)\\
    &Distill    &90.61±0.26&1.91    &0.12±0.08&0.25±0.04&7.42±0.67&0.02±0.01    &0.963±0.005&0.812±0.004    &\textbf{Yes} (10/10)\\
    &\underline{Ours($\mathcal{S}_{LTD}$)} 
        &\underline{90.65±2.12}&\underline{1.70}
        &\underline{\textbf{10.71±0.51}}&\underline{\textbf{13.48±0.08}}
        &\underline{31.23±2.41}&\underline{\textbf{0.40±0.09}}
        &\underline{0.204±0.162}&\underline{0.215±0.053} &\textbf{No} (0/10)\\
    &\underline{Ours($\mathcal{S}_{LSD}$)} 
        &\underline{90.54±2.29}&\underline{1.25}    
        &\underline{6.57±0.31}&\underline{13.42±0.08}
        &\underline{21.61±1.12}&\underline{0.10±0.01}    
        &\underline{\textbf{0.204±0.152}}&\underline{0.173±0.022} &\textbf{No} (0/10)\\
    &Negative   &91.83±0.30&0.62
        &5.89±1.68&7.01±1.36&\textbf{35.92±3.27}&0.29±0.04    
        &0.509±0.123&\textbf{0.016±0.019} &\textbf{No} (0/10)\\
    \cline{2-11}
    & $\tau$ & - & - & 3.78 & 4.78 & 23.92 & 0.20 & 0.652 & 0.405 & - \\
    \hline

    \multirow{7}{*}{\tabincell{c}{\textbf{ImageNet} \\ \textbf{(Acc: 71.13\%)}}}
    &FT(0.5)    &70.17±0.31&0.90    
        &12.36±1.08&0.62±0.15&$\times$&$\times$    &0.982±0.009& 0.815±0.120 &\textbf{Yes} (10/10)\\
    &FT(0.8)    &69.02±0.44&1.93    
        &18.57±2.91&1.70±1.23&$\times$&$\times$    &0.929±0.033&0.766±0.084 &\textbf{Yes} (10/10)\\
    &WP(0.5)    &69.21±0.13&1.81    
        &18.08±3.31&7.34±3.83&$\times$&$\times$    &0.895±0.021&0.641±0.107 &\textbf{Yes} (10/10)\\
    &WP(0.8)    &66.10±1.21&5.25    
        &25.92±1.92&9.30±2.82&$\times$&$\times$    &0.806±0.057&0.579±0.101 &\textbf{Yes} (10/10)\\
    &Distill    &68.29±0.52&2.66    
        &16.28±3.70&3.21±0.43&$\times$&$\times$    &0.994±0.011&0.626±0.062 &\textbf{Yes} (10/10)\\
    &\underline{Ours($\mathcal{S}_{LTD}$)} 
        &\underline{69.39±2.11}&\underline{1.73}
        &\underline{\textbf{142.26±5.69}}&\underline{31.50±2.10}&
        $\times$&$\times$
        &\underline{\textbf{0.248±0.203}} &\underline{0.003±0.0} &\textbf{No} (0/10)\\
    &Negative   &64.57±0.28&6.57
        &134.76±7.58&\textbf{38.87±2.87}&$\times$&$\times$    & 0.353±0.055&\textbf{0.0±0.0} &\textbf{No} (0/10)\\
    \cline{2-11}
    & $\tau$ & - & - & 91.75 & 25.89 & $\times$ & $\times$ & 0.460 & 0.201 & - \\
    \specialrule{1pt}{0pt}{0pt}
\end{tabular}
}
\label{tab:exp11_main_result2}
\end{table*}

In summary, we remove the behavioral patterns of the DNN model at both the latent and logit levels through utilizing a min-max bilevel optimization paradigm. 
This approach allows us to maintain the model's performance even after the removal of fingerprints.

\section{Experiments}
\subsection{Experimental Setup}
We evaluate $\textsc{RemovalNet}$ on five benchmark datasets, i.e.,
  CIFAR10~\cite{krizhevsky2009learning}, GTSRB~\cite{stallkamp2012man}, 
  Skin Lesion Diagnosis~\cite{codella2019skin},
  CelebA~\cite{liu2018large}, and ImageNet~\cite{russakovsky2015imagenet}, covering four scenarios,
  including traffic sign recognition, disease diagnosis, face recognition, and large-scale visual recognition.
Note that we adopt \textbf{``CelebA+20''} and \textbf{``CelebA+31''} to indicate the \textbf{gender} and \textbf{smiling} attributes of the CelebA dataset.
All experiments run on an Ubuntu 20.04 system with 96-core Intel Xeon CPUs and 4 Nvidia GPUs of type GeForce GTX 3090.
In the following context, we describe the generation of the negative, positive, and \textsc{RemovalNet} models. 
Besides, we explain the defense methods and evaluation metrics in our experiments.

\subsubsection{Victim model selection}
We run our experiments on five convolutional neural networks, 
    including ResNet50, VGG19, DenseNet121, InceptionV3, and ViT (ViT\_B/32).
For the input size of $32\times32$, 
    we downsampled the filter size of the first convolutional layer in the original architecture.
For the ImageNet task, the victim models are downloaded from the pre-trained repository of HuggingFace~\footnote{https://github.com/huggingface/pytorch-image-models}.
Table~\ref{tab:exp_setting} summarizes datasets and victim models used in our experiments.
\begin{table}[!ht]
  \centering
  \caption{Datasets and victim models in experiments.}
  \resizebox{0.88\linewidth}{!}{
    \begin{tabular}{lccc}
        \specialrule{1pt}{3pt}{3pt}
        \textbf{Dataset}  & \textbf{Victim model} & \textbf{\#Params} & \textbf{Accuracy(\%)} \\
        \specialrule{1pt}{0pt}{0pt}
        CIFAR10     & VGG19 & 38.96M & 90.58 \\
        \hline
        GTSRB       & InceptionV3 & 21.64M & 92.26 \\
        \hline
        Skin Lesion & ResNet50 & 23.52M & 98.81 \\
        \hline
        CelebA+20/+31   & DenseNet121 & 6.96M & 97.68/92.50 \\
        \hline
        ImageNet    & ViT\_B/32 & 88.19M & 71.13\\
        \specialrule{1pt}{3pt}{3pt}
    \end{tabular}
  }
  \label{tab:exp_setting}
\end{table}
\par\subsubsection{Negative and positive models generation}
Both positive, negative, and \textsc{RemovalNet} models have the same model architecture as the victim model.
We select ten random seeds in all experiments, and the results are averaged over ten random tries.

\begin{figure*}[!t]
  \centering
  \includegraphics[width=0.66\linewidth]{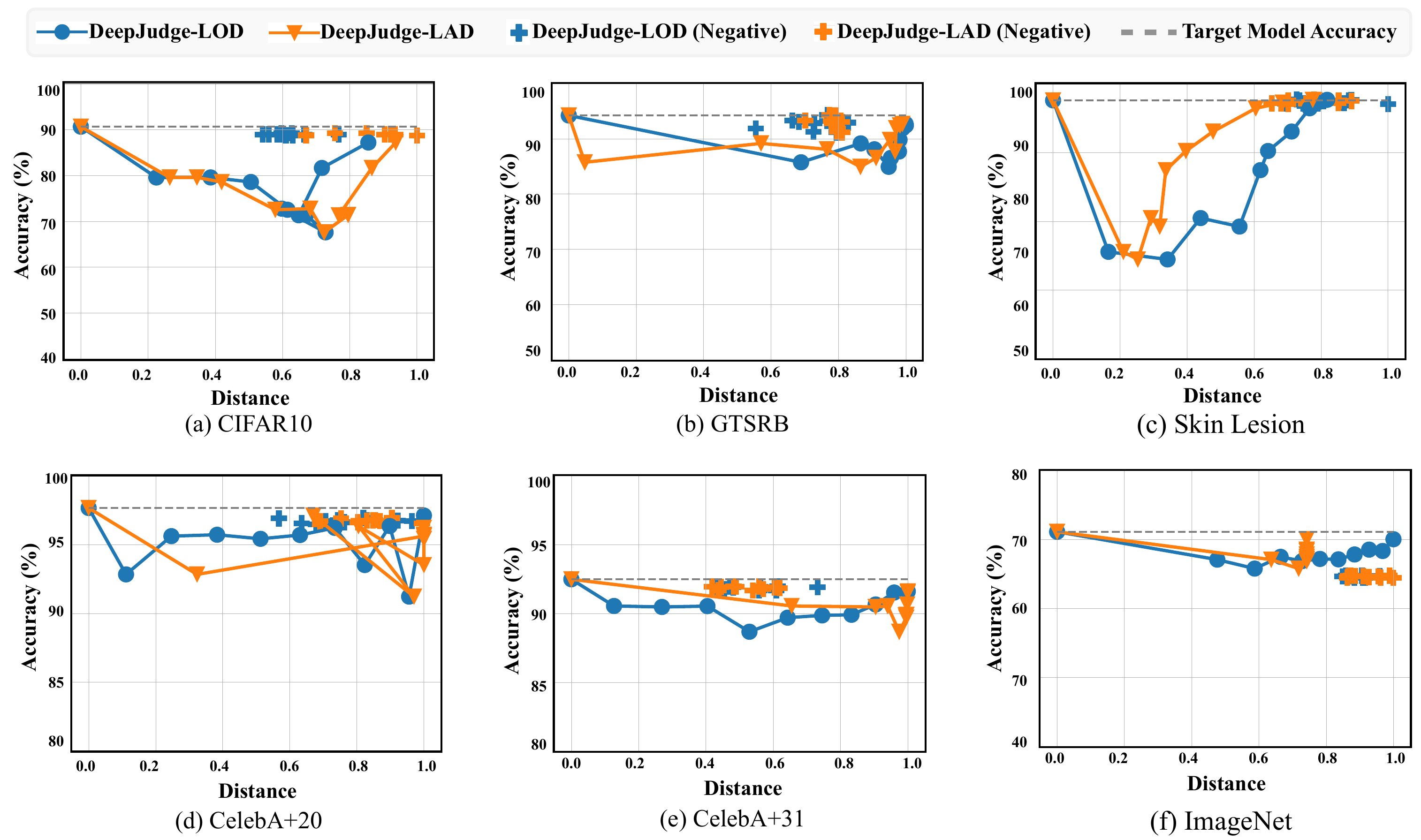}
  \caption{The \textsc{RemovalNet} attack vs. DeepJudge defense.
  Fidelity is presented on the $y$-axis and normalized model distance on the $x$-axis.
  The higher distance and fidelity of the surrogate model, the better performance of our attack.}
  \label{fig:sec5_acc_vs_deepjudge}
\end{figure*}
\par\textbf{Negative models.}
Negative models are trained on the same dataset as the victim model but have different random initializations.
For ImageNet, we conduct our independent training from a pre-trained ``ViT\_B/32\_sam'' model.
In order to simulate a real-world scenario, 
    we randomly add a slight noise to the training iterations and learning rates for each group of random seeds.

\par\textbf{Positive models.}
Four baseline attacks are considered in our experiments, 
    including model fine-tuning, weight pruning, distillation, and model stealing.
Both fine-tuning, weight pruning, and distillation 
    are direct parameters reused from the victim model.
For weight pruning, we first conduct the pruning process 
    and then fine-tune $100\sim500$ iterations on the surrogate model.
For distillation, we distill the knowledge of the victim model using feature distillation.
The model stealing attack serves as a pre-step to the DNN fingerprint removal attack, both of which threaten the copyright of the target model.
We will investigate the efficiency comparison between model stealing attacks and \textsc{RemovalNet} in Section~\ref{sec:exp_computational}.

\textbf{\textsc{RemovalNet}.}
For both scenarios of $\mathcal{S}_{LTD}$ and $\mathcal{S}_{LSD}$, 
    we run our DNN fingerprint removal algorithm with 1000 iterations with a batch size of 128.
The Stochastic Gradient Descent (SGD) optimizer with a momentum of 0.9 is adopted in our method.
The surrogate datasets adopted in $\mathcal{S}_{LSD}$ for the tasks of CIFAR10, CelebA+20, CelebA+31, and Skin Lesion 
    are CINIC10~\cite{darlow2018cinic}, 
    LFW~\cite{huang2008labeled}, VGGFace2~\cite{cao2018vggface2}, and BCN20000~\cite{combalia2019bcn20000}, respectively.
We divide the GTSRB dataset into two non-overlapping partitions and utilize only 15\% of the training set as a surrogate dataset for $\mathcal{S}_{LSD}$.
We set $\alpha=0.2$, $\beta=2.0$, and the learning rate to 0.01 as default hyper-parameters in our experiments.
We will discuss the impact of the substitute dataset's size in Section~\ref{sec:substitute_ratio} 
    and the impact of $\alpha$, $\beta$ in Section~\ref{sec:ablation_study}.

\begin{figure*}[!t]
  \centering
  \includegraphics[width=0.66\linewidth]{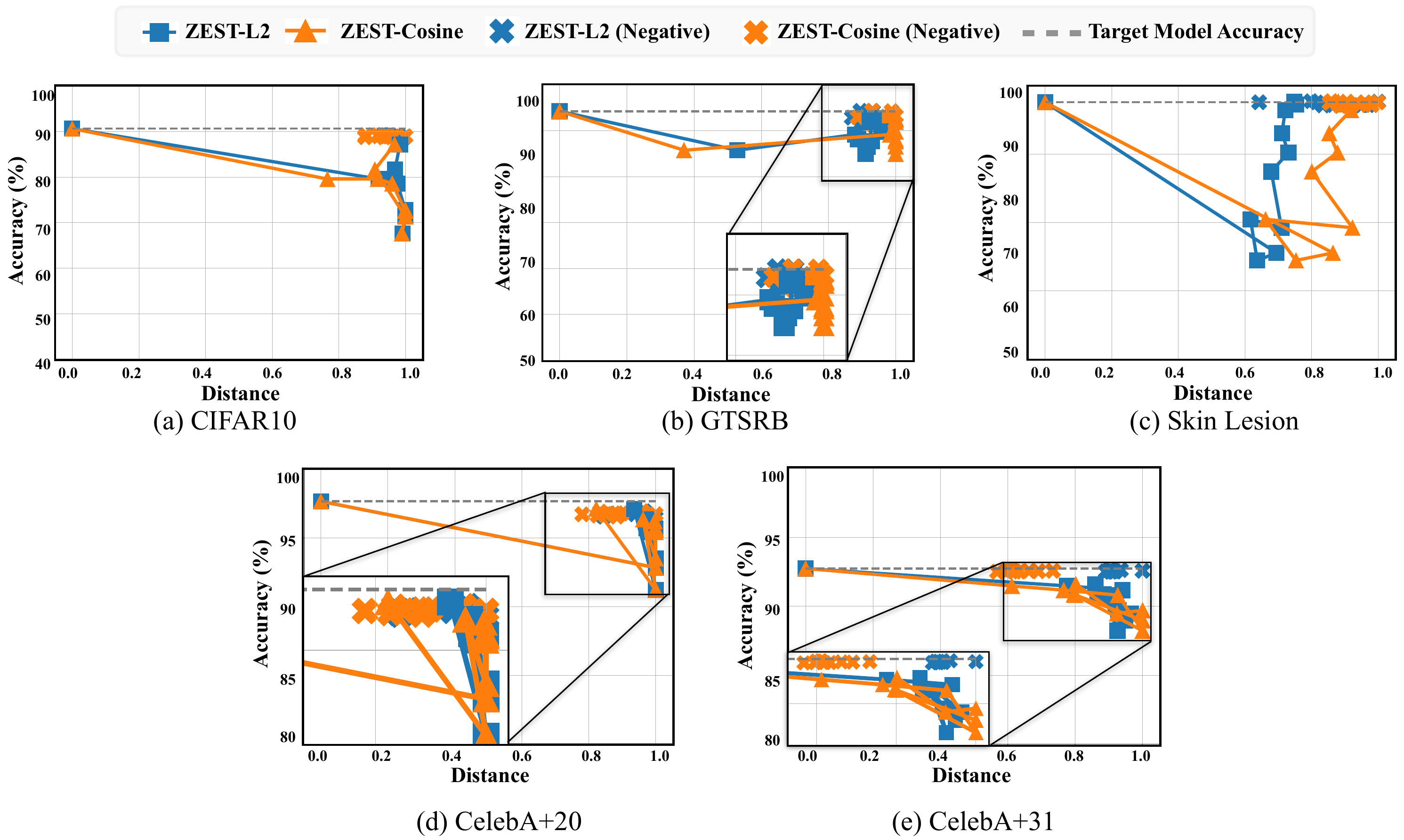}
  \caption{The \textsc{RemovalNet} attack vs. ZEST defense.
  Fidelity is presented on the $y$-axis and normalized model distance on the $x$-axis.
  The higher distance and fidelity of the surrogate model, the better performance of our attack.}
  \label{fig:sec5_acc_vs_zest}
\end{figure*}

\subsubsection{Defense methods and copyright verification metrics}
Four state-of-the-art defense methods (i.e., 
    DeepJudge~\cite{chen2022copy}, ZEST~\cite{jia2021zest}, 
    ModelDiff~\cite{li2021modeldiff},
    and IPGuard~\cite{cao2021ipguard}), 
    including 6+1 metrics, are considered in our experiments.
The evaluation metrics can be categorized into two types: 
    distance (DeepJudge-LOD, DeepJudge-LAD, ZEST-L2, and ZEST-Cosine) 
    and similarity (ModelDiff-DDV and IPGuard-MR).
We have explained the evaluation metrics in Section~\ref{sec:evaluation_metric}.

\subsubsection{Threshold selection}
Due to the differences in statistical characteristics between various datasets and models, 
    it is necessary to calculate the threshold ($\tau$) beforehand. 
Similar to ModelDiff, we select a data-driven model-specific threshold. 
We independently train several reference models that are similar to ${F}_{V}$. 
The threshold can be established as the highest similarity score or lowest distance score among those attained by the reference models.

\subsection{Comparison with Baselines}
In this section, we make a comparison of \textsc{RemovalNet} with baseline attack methods, 
    including fine-tuning, pruning, and distillation on six copyright verification metrics.
We use Table~\ref{tab:exp11_main_result1} and Table~\ref{tab:exp11_main_result2} 
    to evaluate the results of \textsc{RemovalNet} on five benchmark datasets.
In the ``Fidelity'' column, we show the surrogate models' averaged accuracy and accuracy drop.
The columns ``Distance'' and ``Similarity'' illustrate the results of different surrogate models evaluated by various metrics.
FT(0.5) denotes fine-tuning the last 50\% of layers, 
    and WP(0.5) means pruning 50\% weights of the victim model.
It should be noted that large is better for the ``Distance'' column, while small is better for the ``Similarity'' column.
We underline the experimental results of \textsc{RemovalNet} and bold the best results for each metric.
Besides, our goal is not to achieve the best values but to obtain obfuscated results that can evade DNN fingerprinting algorithms.

Overall, the \textsc{RemovalNet} remarkably outperforms all baseline attacks on all six metrics.
Furthermore, the LOD and LAD results of \textsc{RemovalNet} are $\mathbf{\times100}$ \textbf{times} higher 
    than that of the baseline attacks for most experiments, 
    which are very close to the negative models.
For example, in the CIFAR10 dataset, the \textsc{RemovalNet} achieves $9.82\pm0.64$ and $5.13\pm0.27$ (for LOD metric), 
    which is very close to $7.69\pm2.03$ for negative models.
Besides, the \textsc{RemovalNet} also produces deceptive results on ModelDiff and IPGuard.
For example, the \textsc{RemovalNet} achieves $0.159\pm0.125$ and $0.152\pm0.156$ in ModelDiff-DDV for the Skin Lesion dataset, 
    which is deceptive compared with $0.163\pm0.039$ of negative models.
Again, the matching rate (IPGuard) result of \textsc{RemovalNet} on ImageNet is $0.003\pm0.0$, 
    which is deceptive among negative models.
On the other hand, the \textsc{RemovalNet} evades the defense methods by 
    only sacrificing slight fidelity drops.
The most significant fidelity drops are presented 
    in the $\mathcal{S}_{LSD}$ scenario of the Skin Lesion dataset (\textbf{smaller than 4\% drops}).
This phenomenon is because the BCN20000's 
    input distribution and the label space are quite different from the Skin Lesion dataset.
Generally, the \textsc{RemovalNet} can sacrifice negligible accuracy drops 
    to generate a surrogate model that can mislead the ownership verification algorithms.

\subsection{Fidelity and Effectiveness of \textsc{RemovalNet}}
\label{sec:exp_effectiveness}
The trade-off between fidelity and effectiveness is a great challenge of DNN fingerprint removal attacks.
In this section, we show the learning process of \textsc{RemovalNet} to find the optimal surrogate model, 
    which achieves high fidelity while removing the victim model’s behavioral patterns. 
Figure~\ref{fig:sec5_acc_vs_deepjudge} and Figure~\ref{fig:sec5_acc_vs_zest} depict the trade-off between the fidelity 
    and attack performance of \textsc{RemovalNet}.
We sample every 100 iterations for \textsc{RemovalNet}, with 11 checkpoints (the first checkpoint is $t=0$).
As illustrated in Figure~\ref{fig:sec5_acc_vs_deepjudge} and Figure~\ref{fig:sec5_acc_vs_zest}, 
    the fidelity of the \textsc{RemovalNet} first experiences a rapid decline, 
    then progressively grows and finally reaches a stable value.
The reason behind the fidelity rapid decline phenomenon is 
    that the removal procedure over the latent-level and logit-level introduces the side effect of catastrophic forgetting.
On the other hand, the results of DeepJudge and ZEST experienced a violent fluctuation during this period and finally reached a high value.
Since the \textsc{RemovalNet} does not know the DNN fingerprinting strategy and the fingerprint set, 
    removing the behavioral patterns of the victim model is an indirect optimization process.
As a result, the results of DeepJudge and ZEST fluctuate during this period.
Although the fidelity of \textsc{RemovalNet} has slight drops, 
    we can sacrifice only a small fidelity to bypass the defense methods. 
In conclusion, the \textsc{RemovalNet} attack is effective 
    in removing the victim model’s behavioral patterns and evading defense methods, only sacrificing little fidelity.

\subsection{Efficiency of \textsc{RemovalNet}}
\label{sec:exp_efficiency}
The efficiency of the DNN fingerprint removal attack contains two aspects: 
    substitute set’s size and computing resources (Section~\ref{sec:threat_model}).

\begin{figure}[!h]
  \centering
  \includegraphics[width=\linewidth]{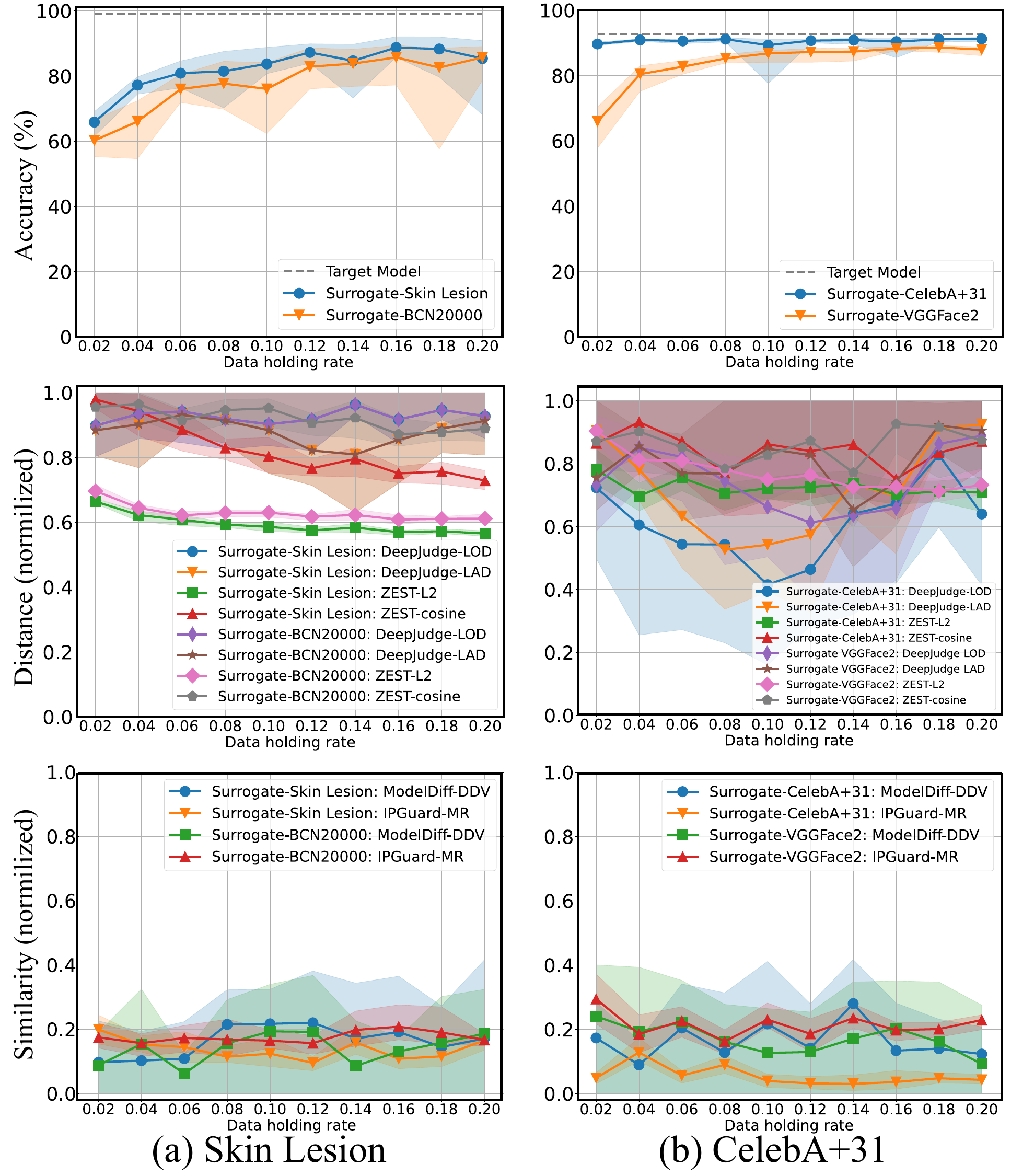}
  \caption{The performance of \textsc{RemovalNet} with different ratios of substitute data. 
        The first to third rows show the fidelity, distance, and similarity that vary with the data holding rate.}
  \label{fig:sec5_exp31_efficiency}
\end{figure}
\subsubsection{Substitute set’s size.}
\label{sec:substitute_ratio}
In confidential scenarios (e.g., disease diagnosis and facial recognition), 
    datasets are collected and protected by organizations.
In this practical context, 
    the substitute set’s size is critical to the utility of DNN fingerprint attacks.

For both $\mathcal{S}_{LTD}$ and $\mathcal{S}_{LSD}$ scenarios,
    we evaluate the performance of \textsc{RemovalNet} under different ratios of substitute data (from 2\% to 20\%).
In order to avoid catastrophic forgetting, we decrease the learning rate in this experiment.
As illustrated in Figure~\ref{fig:sec5_exp31_efficiency}, 
    the \textsc{RemovalNet} achieves higher fidelity with higher data holding ratios.
However, the distance metric slightly decreases with higher data-holding ratios.
This phenomenon is reasonable and within expectation.
Due to a lower learning rate, the optimizer adjusts fewer weights, making it more challenging to remove the behavioral patterns remaining in the victim model.
It should be noted that for the Skin Lesions and VGGFace2 (p=2\%) dataset (after Random Over-Sampling Examples, ROSE~\cite{menardi2014training} labels rebalance), 
    the adversary only holds \textbf{750 and 400 samples} (\textbf{saving 92.51\% and 99.80\% training data}) to conduct the DNN fingerprint removal attack, respectively.
Overall, even with a small ratio of substitute data, the \textsc{RemovalNet} still has comparable fidelity and ownership verification evading capability.

\begin{figure}[!t]
  \centering
  \includegraphics[width=0.95\linewidth]{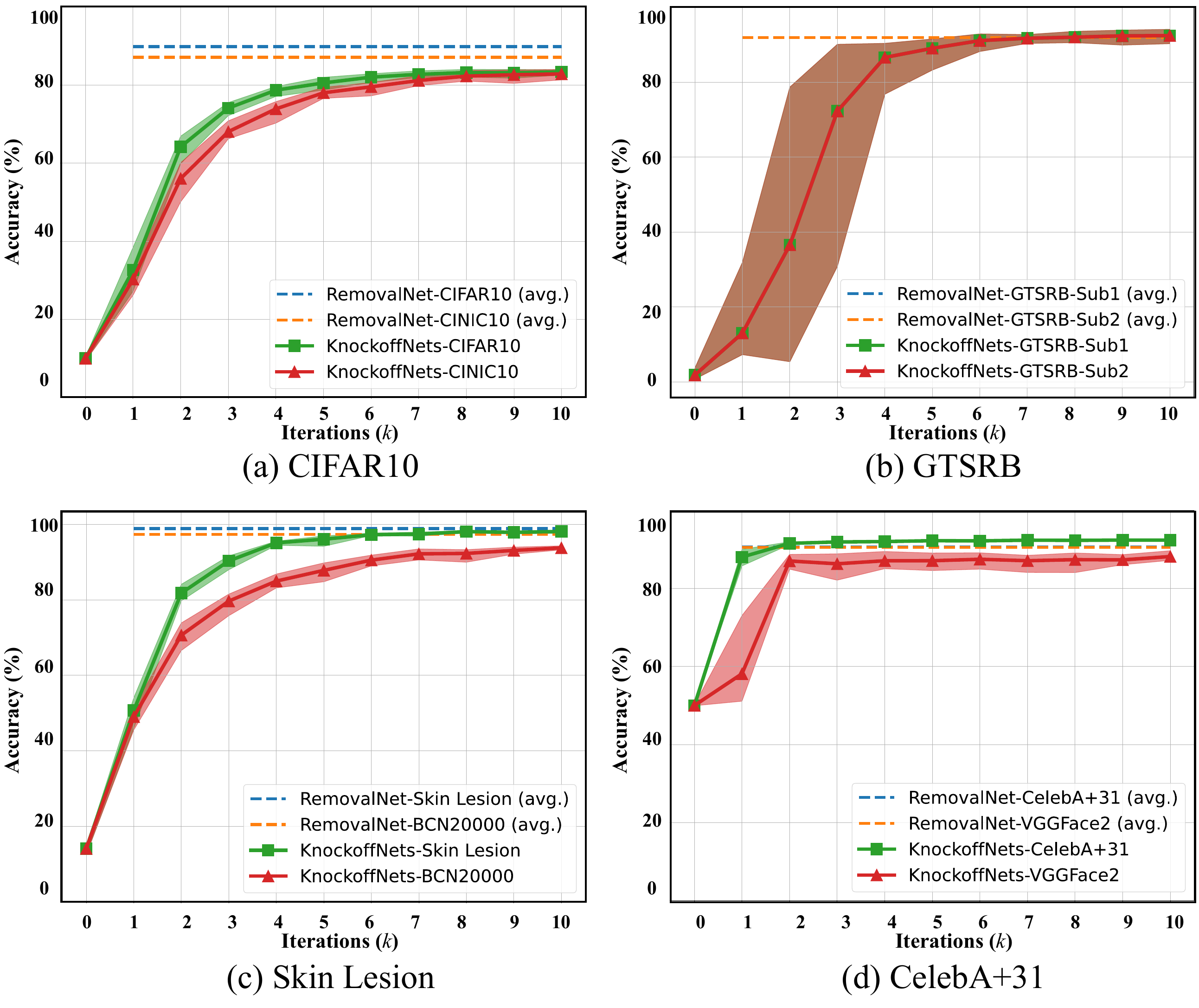}
  \caption{The comparison of the computational efficiency between \textsc{RemovalNet} (dashed line) and model stealing attacks (solid line).}
  \label{fig:sec5_exp32_efficiency_computation}
\end{figure}
\subsubsection{Computational resources}
\label{sec:exp_computational}
Until recently, no DNN fingerprint removal attack has been comparable to \textsc{RemovalNet} in terms of effectiveness.
Therefore, we have decided to use the model stealing attack as a benchmark to evaluate the computational resources required by \textsc{RemovalNet}.
In this context, we make a comparison of computation efficiency between \textsc{RemovalNet} and model stealing attacks.

KnockoffNets~\cite{orekondy2019knockoff} is an efficient model stealing attack 
    that exploits the active learning paradigm to increase stealing efficiency.
Similar to \textsc{RemovalNet}, we consider $\mathcal{S}_{LTD}$ and $\mathcal{S}_{LSD}$ scenarios for KnockoffNets 
    and set the query budge to $batch\_size \times 10000$ (10,000 iterations).
Since there is no substitute dataset comparable to ImageNet, 
    we only evaluate the efficiency between \textsc{RemovalNet} and KnockoffNets on the CIFAR10, GTSRB, Skin Lesion, and CelebA.
Figure~\ref{fig:sec5_exp32_efficiency_computation} depicts the learning curves of KnockoffNets (solid green and red lines) 
    and the average accuracies of \textsc{RemovalNet} (dashed blue and orange lines) on VGG19.
We can observe from Figure~\ref{fig:sec5_exp32_efficiency_computation} that, in the CelebA+31 dataset, 
    KnockoffNets uses nearly $2000\sim3000$ iterations to achieve a fidelity comparable to \textsc{RemovalNet} 
    ($\mathbf{\times2}$ \textbf{times} higher than \textsc{RemovalNet}).
Even worse, the training overhead of KnockoffNets reaches 
    $6000\sim8000$ iterations in the CIFAR10, 
    which is nearly $\mathbf{\times7}$ \textbf{times} higher than \textsc{RemovalNet}.
Overall, the \textsc{RemovalNet} is computationally efficient in saving $50\sim85$\% of computational resources compared with advanced model stealing attacks.

\subsection{Ablation Study}
\label{sec:ablation_study}
In this section, we discuss two impacts of \textsc{RemovalNet} (i.e., the impact of latent-level and logit-level removal) and explain two phenomena of \textsc{RemovalNet}, including the visualization of the activated feature maps and decision boundary.

\begin{figure}[!t]
  \centering
  \includegraphics[width=0.95\linewidth]{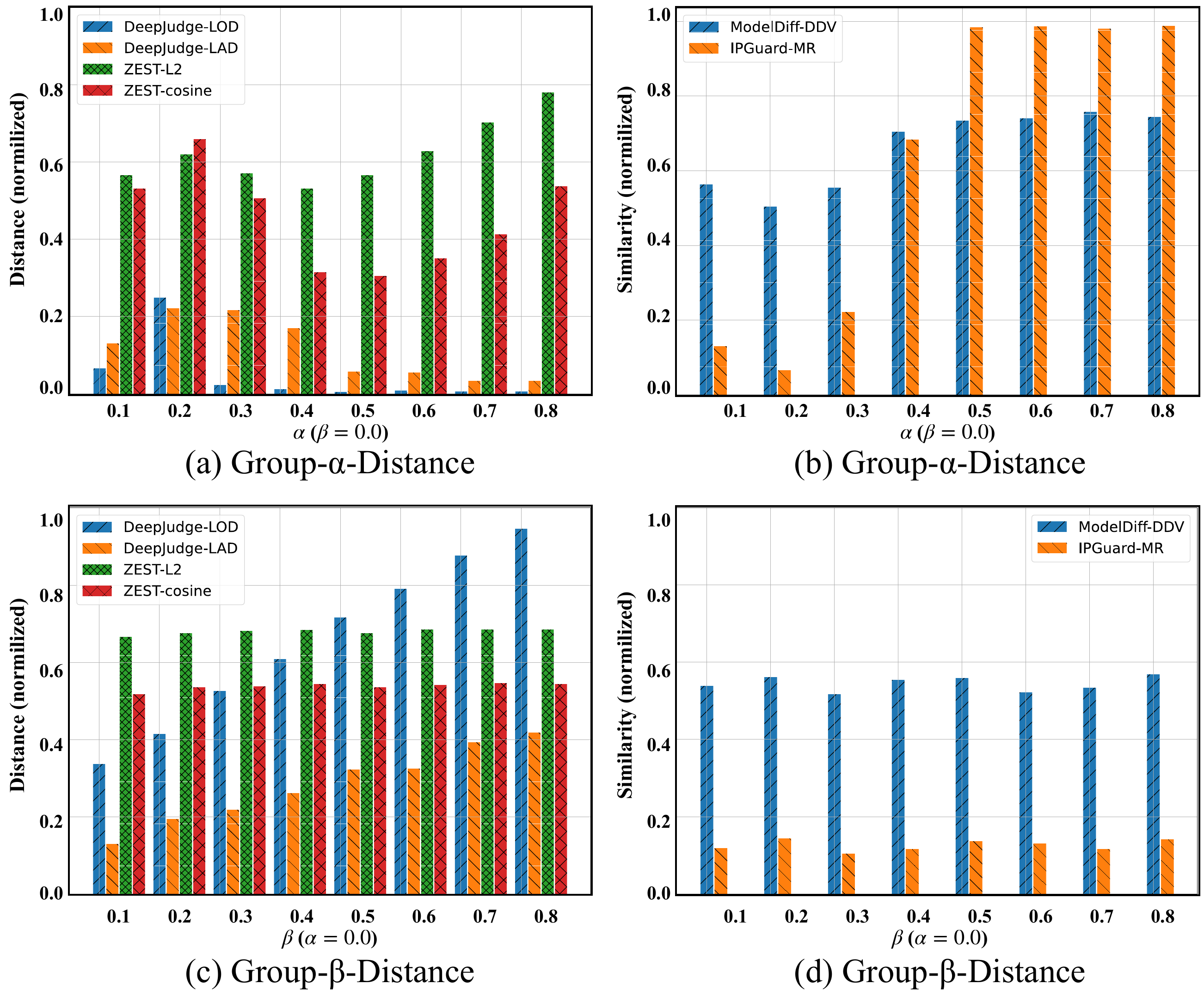}
  \caption{Performance change of \textsc{RemovalNet} with different ratios of $\alpha$ and $\beta$ on CIFAFR10.
  The results are averaged with ten created surrogate models.}
  \label{fig:sec5_exp41_alpha_beta}
\end{figure}
\subsubsection{Impacts of the latent-level and logit-level removal.}
As mentioned in Section~\ref{sec:method}, the DNN fingerprint removal procedure is conducted on the latent-level and logit-level, 
    which are controlled by $\alpha$ and $\beta$. 
Specifically, the $1-\alpha$ controls the KL-divergence loss, 
    $\alpha$ controls the cross-entropy loss, and the $\beta$ controls the $\mathcal{L}_{feat}$.
In this experiment, we set the learning rate to 0.03. 
When evaluating the impact of $\alpha$, we set $\beta=0$, and vice versa.
Figure~\ref{fig:sec5_exp41_alpha_beta} (a) and (b) depict the distance and similarity score of \textsc{RemovalNet} with various $\alpha$.
We can observe from subfigures (a) and (b) that with the $\alpha$ increasing, the ZEST, ModelDiff, and IPGuard gradually grow.
In contrast, the influence of $\alpha$ on DeepJudge's performance remains relatively minor.
The experimental results indicate that large $\alpha$ is better for ZEST, while small $\alpha$ is better for ModelDiff and IPGuard.
This is because a larger cross-entropy constraint leads to greater similarity between the surrogate and victim models within the decision boundary.
Figure~\ref{fig:sec5_exp41_alpha_beta} (c) and (d) depict the performance of \textsc{RemovalNet} with various $\beta$.
We can observe from subfigures (c) and (d) that the results of \textsc{RemovalNet} on DeepJudge progressively rise with the growing $\beta$.
In contrast, the impact of $\beta$ on the performance of ZEST, ModelDiff, and IPGuard remains comparatively negligible.
This is because the influence of latent-level removal on the decision boundary is small, leading to only small changes in performance on ZEST, ModelDiff, and IPGuard.

\subsubsection{Visualization of saliency maps.}
\begin{figure}[!t]
  \centering
  \includegraphics[width=0.85\linewidth]{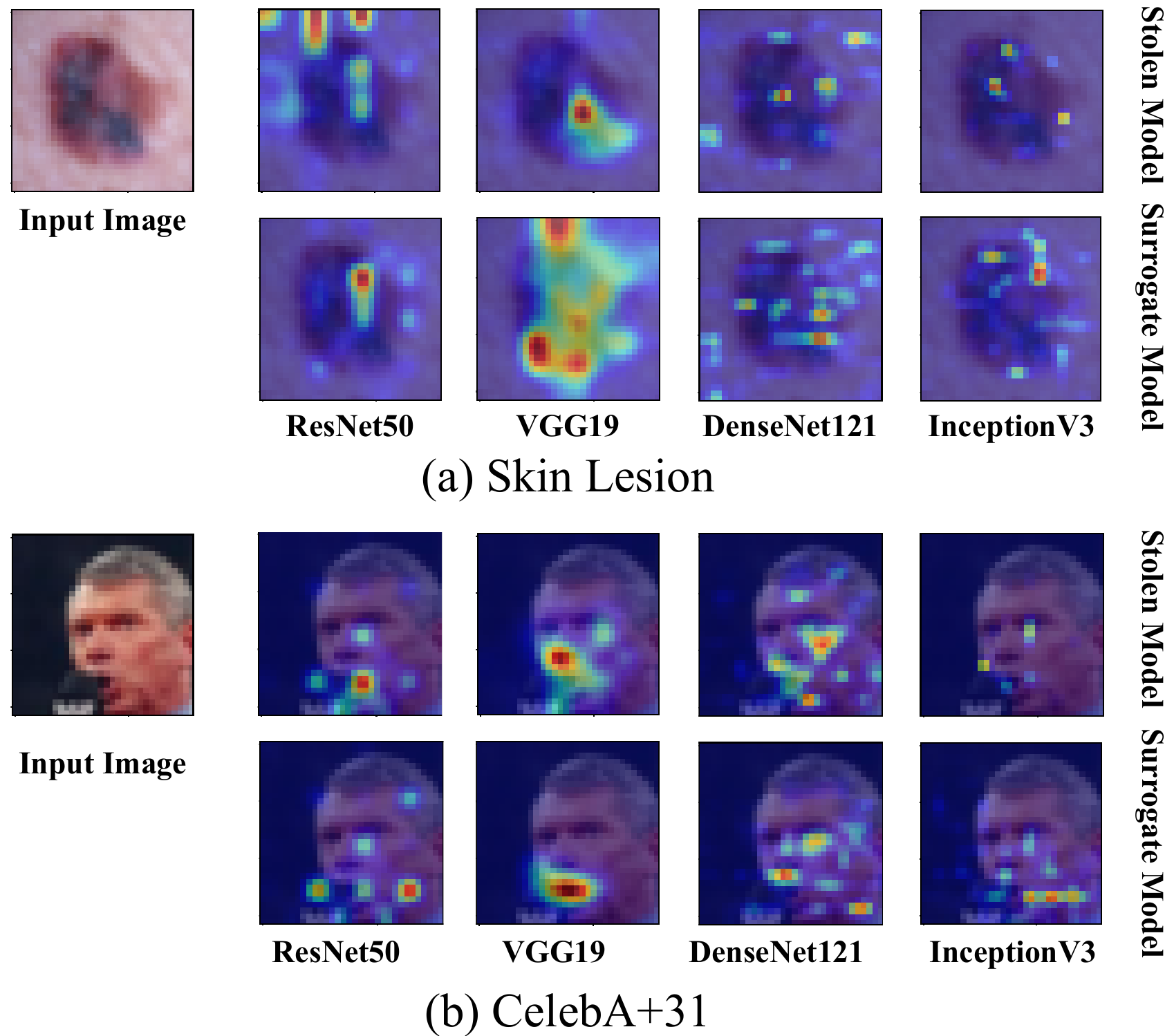}
  \caption{Examples of saliency maps for Layer\#2 of the victim model and surrogate model generated by \textit{LayerCAM}~\cite{jiang2021layercam}.}
  \label{fig:sec5_saliency_maps}
\end{figure}
The saliency maps provide a technique to explore the hidden layers of CNN, 
    which can be used to interpret the behavior of a deep neural network. 
In this section, we employ the \textit{LayerCAM}~\cite{jiang2021layercam} to produce reliable class activation maps 
    (i.e., the saliency maps) to show the behavioral patterns changes of the surrogate model in the hidden layers.
To facilitate a more comprehensive comparison, we attack four distinct network architectures to demonstrate variations in saliency maps.
Figure~\ref{fig:sec5_saliency_maps} exhibits the heatmap visualization of 
    the class activation maps on the Skin Lesion and CelebA+31 datasets.
In the first and second rows of Figure~\ref{fig:sec5_saliency_maps}, 
    the heatmap highlights the activated areas of the victim and surrogate models.
We can observe that the activated areas are remarkably different between the target and surrogate models.
This phenomenon is because the min-max bilevel optimization distills the knowledge of the victim model, 
    accompanied by maximizing the distance of feature maps between the victim model and surrogate models.
In general, the \textsc{RemovalNet} is capable of removing the behavioral patterns 
    in the latent representations by activating different neurons.

\subsubsection{Visualization of the decision boundary.}
\begin{figure}[!t]
  \centering
  \includegraphics[width=\linewidth]{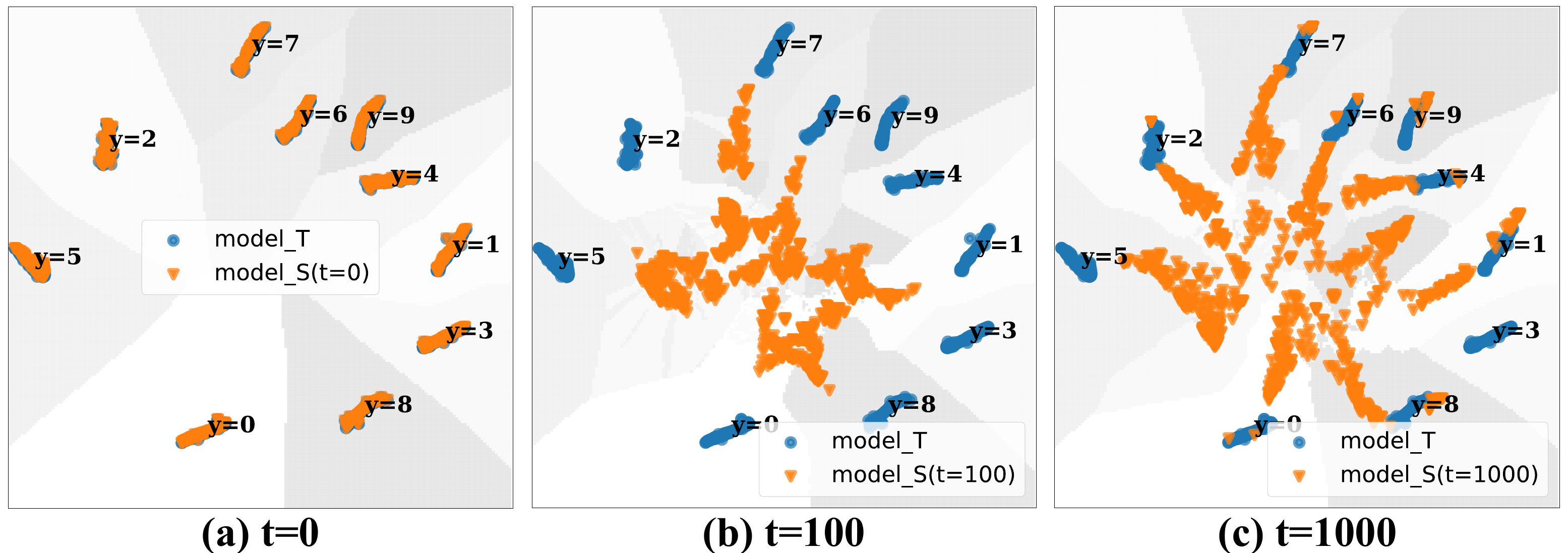}
  \caption{Decision boundary changes of the surrogate model ${F}_{S}$ on CIFAR10.}
  \label{fig:sec5_dcb_changes}
\end{figure}

Figure~\ref{fig:sec5_dcb_changes} depicts the decision boundary changes of 
    the surrogate model of \textsc{RemovalNet} trained on CIFAR10.
We randomly select 100 test samples for each class and 
    exploit T-SNE to embed the output probabilities of the surrogate and victim models.
We can observe from Figure~\ref{fig:sec5_dcb_changes} 
    that the outputs of the surrogate model overlap with that of the victim model at the beginning.
After 100 iteration optimization, 
    the outputs of the surrogate model move into the center of the coordinate, 
    and the output difference becomes obfuscated.
Finally, after 1000 iterations, the decision boundary of the surrogate model 
    becomes significantly different from the victim model.
Figure~\ref{fig:sec5_dcb_changes} demonstrates that the DNN fingerprint removal over the logit-level 
    changes the behavioral patterns of the victim model over the decision boundary.

\section{Discussions}
\subsection{Possible Countermeasures}
\partitle{Adversarial attacks~\cite{chen2022copy}}
Chen \textit{et al.} propose a model distance metric, called \textit{Robustness Distance (RobD)},
    to measure the robustness distance between the victim and suspected models~\cite{chen2022copy}.
RobD’s core insight is that a model’s robustness (Rob) is linked to the decision boundary,
    while the decision boundary is learned by a unique optimization process.
Generally, robustness is a property of DNN, which comes from the non-robustness features in training data~\cite{ilyas2019adversarial}. 
Therefore, injecting non-robustness features during training can promote the robustness distance between the victim and surrogate models.
In fact, the removal process introduces noise variation into the latent representations, which are non-robust features.
However, injecting excessive non-robust features will inevitably injure the robustness of the surrogate model.
Ultimately, our main concern should be the adverse impact of DNN fingerprint removal, 
    which can make the surrogate model more vulnerable to adversarial attacks.

\begin{table}[!h]
\centering
\caption{Performance of the potential countermeasure on CIFAR10. $RobD_{{F}_{S}}$ denotes the RobD between ${F}_{V}$ and ${F}_{S}$.}
\resizebox{\linewidth}{!}{
\begin{tabular}{l|c|c|c|c|c}
    \specialrule{1pt}{0pt}{0pt}
    \diagbox{Method}{Metric} & $Rob_{{F}_{V}}$ & $Rob_{{F}_{S}}$ & $Rob_{Neg}$ & $RobD_{{F}_{S}}$ & $RobD_{Neg}$ \\
    \specialrule{1pt}{0pt}{0pt}
    FGSM & 0.694 & 0.553 & 0.599 & 0.141 & 0.095 \\ \hline
    PGD & 0.501 & 0.422 & 0.419 & $\underline{0.079}$ & $\underline{0.083}$ \\ \hline
    DeepFool & $\underline{0.070}$ & $\underline{0.171}$ & 0.121 & 0.101 & 0.051 \\
    \specialrule{1pt}{0pt}{0pt}
\end{tabular}}
\label{fig:adv_atk}
\end{table}

\partitle{Evaluation}
In this context, we use Rob and RobD metrics to investigate the feasibility of the adversarial attacks on the CIFAR10 dataset.
The Rob is defined as its fidelity on adversarial set $\mathcal{D}_{adv}$:
\begin{equation}
    Rob({F}_{V}, \mathcal{D}_{adv}) = 
        \frac{1}{|\mathcal{D}_{adv}|} 
            \sum_{(x_{i}, y_{i}) \in \mathcal{D}_{adv}}{\mathbbm{1}[{F}_{V}(x_{i}) = y_{i}]}
\end{equation}
Furthermore, the robustness distance between the victim and surrogate models is defined as:
\begin{small}
\begin{gather}
    RobD({F}_{V}, {F}_{S}, \mathcal{D}_{adv}) = 
        |Rob({F}_{V}, \mathcal{D}_{adv}) - Rob({F}_{V}, \mathcal{D}_{adv})|
\end{gather}
\end{small}
We report the results of Rob and RobD against FGSM, PGD, and DeepFool in Table~\ref{fig:adv_atk}.
As illustrated in Table~\ref{fig:adv_atk},
(1) both the target, surrogate, and negative models are not robust to adversarial attacks,
(2) the surrogate model's robustness drops in FGSM and PGD but has a small increase in DeepFool (i.e., $Rob_{{F}_{V}}=0.07$ vs $Rob_{{F}_{S}}=0.171$),
and (3) for the PGD attack, $RobD_{{F}_{S}}=0.079$ is close to $RobD_{{F}_{S}}=0.083$, 
    meaning it's difficult only to use RobD to determine whether a suspected model is a copy of the victim model.

\subsection{Limitations and Future Work}
\textbf{Evaluation on various tasks.}
In this paper, we concentrate on computer vision tasks.
At the same time, we acknowledge that models from various domains, such as natural language processing and graph neural networks, also hold significant privacy and copyright value.
Notably, the transformers of pre-trained language models (e.g., LLaMA and GPT) possess a considerable capacity for embedding non-robustness features. 
Moreover, the powerful feature extractor of graph neural networks also can be used to remove fingerprint-specific knowledge.
As such, we believe that the \textsc{RemovalNet} can be adapted to these tasks. 
We plan to extend \textsc{RemovalNet} to language and graph models in future work.

\section{Related Works}
\subsection{DNN Fingerprint Removal Attacks}
Despite the availability of general-purpose schemes, there is currently no effective attack specifically designed for removing DNN fingerprints. 
In this paper, we consider the model fine-tuning, weight pruning, and distillation as the baseline of our method. 
Furthermore, we discuss the distinctions between DNN fingerprint removal and model stealing attacks.

\textbf{Model fine-tuning.}
Model fine-tuning~\cite{guo2019spottune} is a machine-learning method that transfers the knowledge of a victim model to a surrogate model.
The most common way of fine-tuning is to reuse the feature extraction module of the victim model and retrain the classification layers.

\textbf{Model compression.}
Model pruning~\cite{zhuang2019rethinking} is a model compression technique 
    that discards the model weights while maintaining its classification performance.
The typical model pruning algorithm includes three phases: 
    model training, weight pruning, and fine-tuning.
Through specific criteria, the redundant weights are pruned, while the valuable weights will be kept to maintain the model performance.
Knowledge distillation~\cite{gou2021knowledge} is another model compression technique 
    that transfers knowledge from a victim model to create a surrogate model.

\textbf{Model stealing.}
In model stealing attacks, the adversary queries the victim model using substitute data 
    and leverages the data labeled by the victim model to create his surrogate model.
JBA~\cite{papernot2017practical} and Knockoff~\cite{orekondy2019knockoff} are generic model stealing attacks.

\subsection{DNN Fingerprinting} 
DNN fingerprinting is an ownership verification technique that protects the intellectual property of the DNN model owner.
DNN fingerprinting relies on the intrinsic characteristics of the neural networks
to verify whether a suspected model is a copy of the victim model.
Those characteristics can be represented by behavioral patterns in the latent representations and decision boundaries.

Recent works focus on investigating many metrics to measure 
    the distance or similarity between the victim and suspected models.
Existing works on DNN fingerprinting can be categorized into 
    the white-box~\cite{chen2022copy}
    and black-box methods
    ~\cite{cao2021ipguard,jia2021zest,li2021modeldiff,pan2022metav,yang2022metafinger,lukas2021deep,pratyush2021dataset,zhang2022remos}.
For the white-box methods, the verifier is assumed to access the intermediate layers of the suspected model.
The authors of DeepJudge~\cite{chen2022copy} propose multiple metrics 
    (e.g., Layer Outputs Distance, Layer Activation Distance) 
    to measure the distance of intermediate layers between the victim and suspected models.
On the other hand, in black-box methods, 
    the verifier can only query the suspected model with probing samples and observe its posterior probabilities.
For example, the authors of~\cite{cao2021ipguard} propose IPGuard, 
    which extracts plenty of adversarial examples near the decision boundary of the victim model.
Compared with the victim model, 
    once the suspected model has similar responses to the extracted adversarial examples (i.e., matching rate), 
    the suspected model is determined as a copy of the victim model.
The authors of~\cite{li2021modeldiff} propose ModelDiff, 
    which inspects the behavioral patterns of a suspected model using the DDV.
The ModelDiff outputs a similarity score to measure the behavioral patterns similarity of two models over the decision boundary.
Besides, the authors of ~\cite{peng2022fingerprinting} adopt 
    Universal Adversarial Perturbations (UAPs) as fingerprints to verify the copyright of the DNN model.
Different from the aforementioned methods, the authors of ~\cite{lukas2021deep,yang2022metafinger,pan2022metav} first generate many negative models and positive models, 
    and then employ those models as training data to generate adversarial examples.
The objective of those methods is to find conferrable adversarial examples 
    that exclusively transfer a target label from the victim model to the surrogate model.
Finally, the fidelity of the suspected model on those adversarial examples is used to determine 
    whether the suspected model is a copy of the victim model.

\section{Conclusions}
\label{sec:conclusion}
This paper presents the first comprehensive investigation of DNN fingerprint removal attacks.
We provided analytical and empirical evidence for the feasibility of the DNN fingerprint removal attack.
In particular, we propose a bilevel optimization-based DNN fingerprint removal attack named \textsc{RemovalNet}.
We conduct extensive experiments to evaluate the \textbf{fidelity}, \textbf{effectiveness}, and \textbf{efficiency} of \textsc{RemovalNet}, 
    covering five benchmark datasets and four diverse scenarios, namely, traffic sign recognition, disease diagnosis, face recognition, and large-scale visual recognition.
Our findings reveal significant threats to the verification of DNN ownership 
    and highlight the urgent need for robust copyright protection methods, particularly against DNN fingerprint removal attacks.
We hope this work can serve as a wake-up call to the dangers posed by these types of attacks 
    and motivate the scientific community to develop robust measures to protect against them.

{
  \small
  \bibliographystyle{IEEEtran}
  \bibliography{main}
}
\input{static/bio/bio}
\end{document}

%% file: static/bio/bio.tex

\begin{IEEEbiography}[{\includegraphics[width=1in,height=1.25in,clip,keepaspectratio]{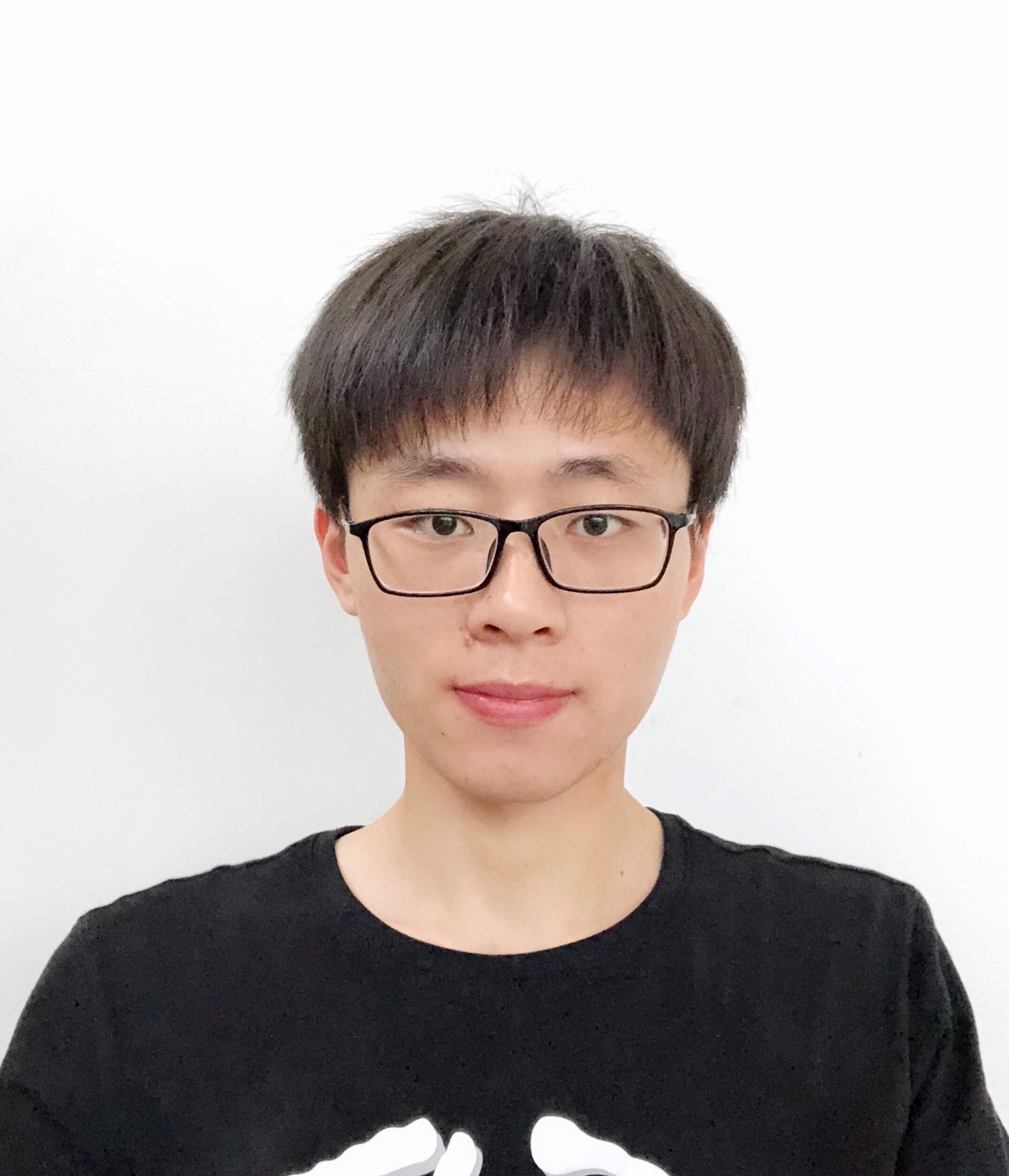}}]{Hongwei~Yao}
is currently a Ph.D. student at the College of Computer Science and Technology, Zhejiang University , advised by Dr. Zhan Qin.
Prior to that, he obtained his bachelor (2016) and master (2020) degrees from Hangzhou Dianzi University. His research focuses on machine learning security and privacy.
\end{IEEEbiography}
\begin{IEEEbiography}[{\includegraphics[width=1in,height=1.25in,clip,keepaspectratio]{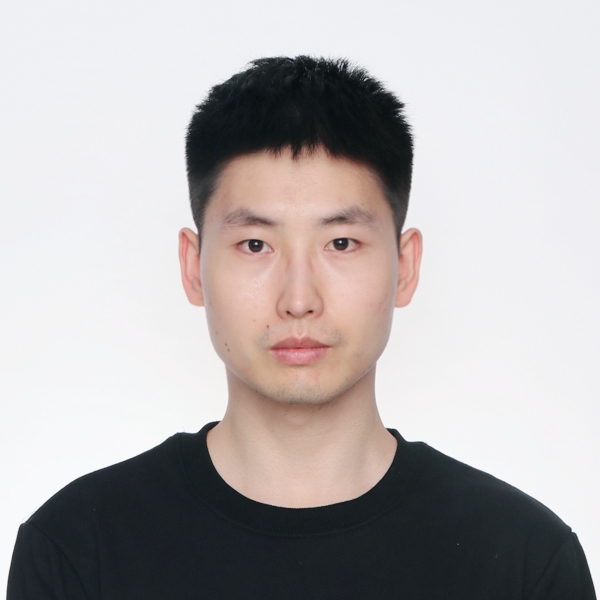}}]{Zheng~Li}
is currently a Ph.D. student at CISPA Helmholtz Center for Information Security, advised by Dr. Yang Zhang. Prior to that, he obtained his bachelor (2017) and master (2020) degrees from Shandong University under the supervision of Prof. Shanqing Guo. His research focuses on machine learning security and privacy.
\end{IEEEbiography}
\begin{IEEEbiography}[{\includegraphics[width=1in,height=1.25in,clip,keepaspectratio]{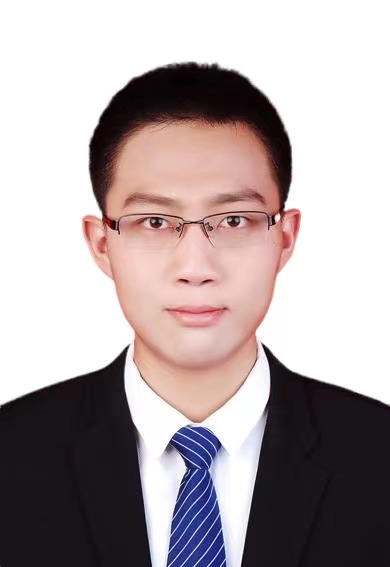}}]{Kunzhe Huang}
received the B.S. degree from the Department of Computer Science and Engineering, Central South University, in 2019. He is a graduate student at the Department of Computer Science and Technology, Zhejiang University. His research interests include security and privacy in deep learning.
\end{IEEEbiography}
\begin{IEEEbiography}[{\includegraphics[width=1in,height=1.25in,clip,keepaspectratio]{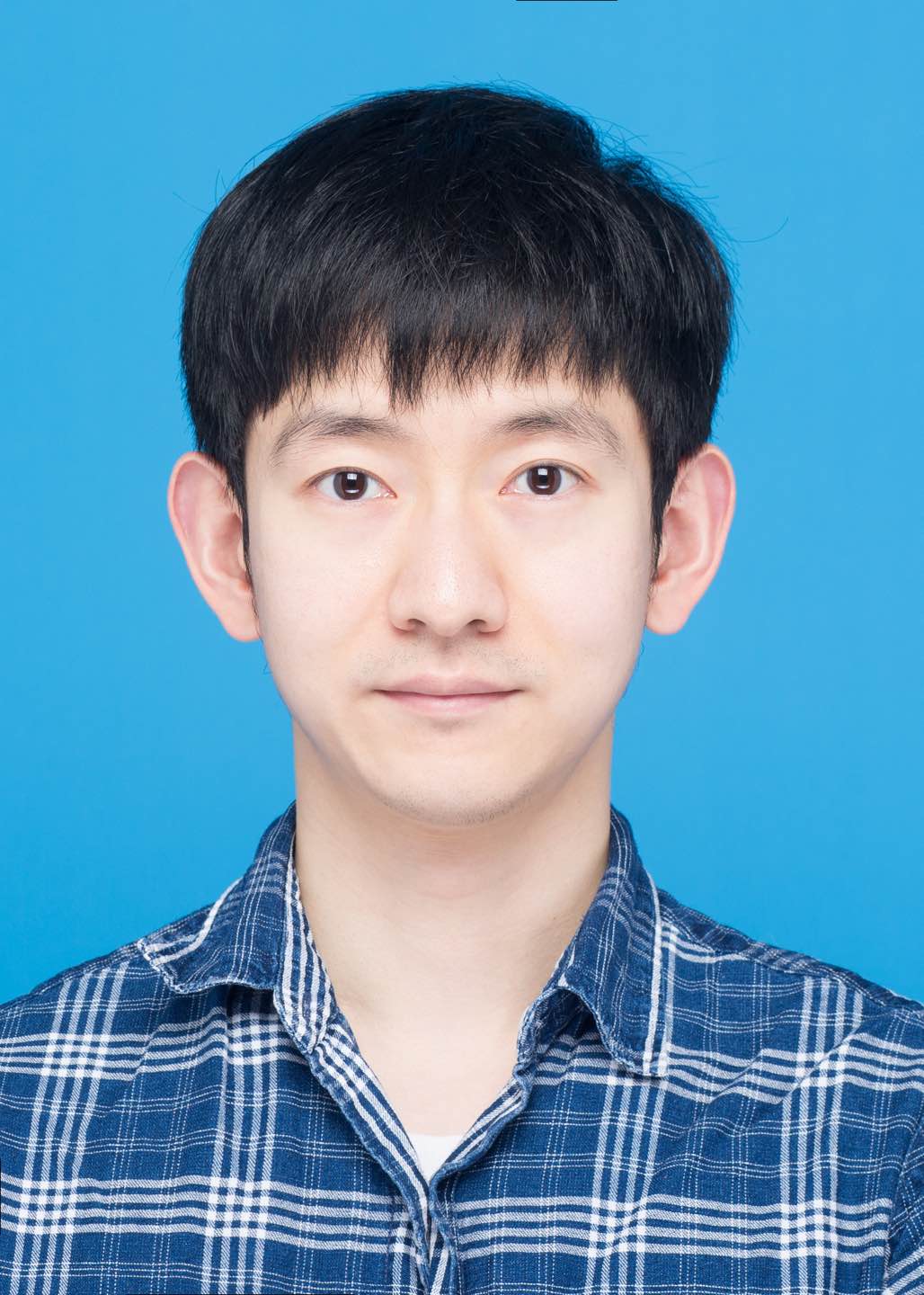}}]{Jian Lou}
is currently a researcher at Hangzhou Innovation Center at Zhejiang University. He was an associate professor at Xidian University from 2021 to 2022. He was a Post-doc in the Department of Computer Science at Emory University from 2019 to 2021. He obtained Ph.D. in Computer Science at Hong Kong Baptist University in 2018. Prior to that, He received a B.S. in Mathematics from Zhejiang University in 2013. 
\end{IEEEbiography}
\begin{IEEEbiography}[{\includegraphics[width=1in,height=1.25in,clip,keepaspectratio]{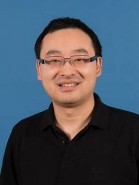}}]{Zhan Qin}
is currently a ZJU100 Young Professor, with both the College of Computer Science and Technology and the Institute of Cyberspace Research (ICSR) at Zhejiang University, China. He was an assistant professor at the Department of Electrical and Computer Engineering in the University of Texas at San Antonio after receiving the Ph.D. degree from the Computer Science and Engineering department at State University of New York at Buffalo in 2017. His current research interests include data security and privacy, secure computation outsourcing, artificial intelligence security, and cyber-physical security in the context of the Internet of Things. His works explore and develop novel security sensitive algorithms and protocols for computation and communication on the general context of Cloud and Internet devices.
\end{IEEEbiography}
\begin{IEEEbiography}[{\includegraphics[width=1in,height=1.25in,clip,keepaspectratio]{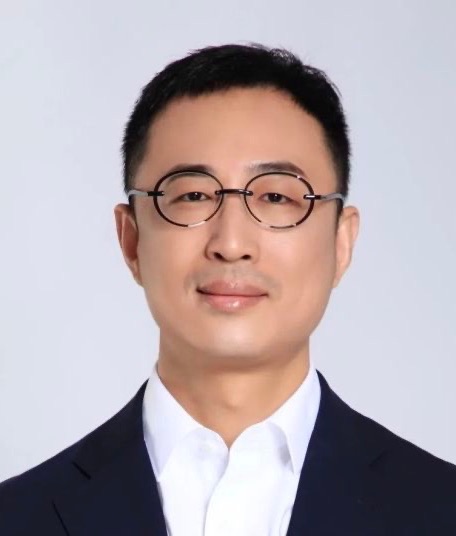}}]{Kui Ren}
is a Professor and the Dean of School of Cyber Science and Technology at Zhejiang University. Before that, he was SUNY Empire Innovation Professor at State University of New York at Buffalo. He received his PhD degree in Electrical and Computer Engineering from Worcester Polytechnic Institute. Kui’s current research interests include Data Security, IoT Security, AI Security, and Privacy. He received Guohua Distinguished Scholar Award from ZJU in 2020, IEEE CISTC Technical Recognition
Award in 2017, SUNY Chancellor’s Research Excellence Award in 2017, Sigma Xi Research Excellence Award in 2012 and NSF CAREER Award in 2011. Kui has published extensively in peer-reviewed journals and conferences and received the Test-of-time Paper Award from IEEE INFOCOM and many Best Paper Awards from IEEE and ACM including MobiSys’20, ICDCS’20, Globecom’19, ASIACCS’18, ICDCS’17, etc. His h-index is 74, and his total publication citation exceeds 32,000 according to Google Scholar. He is a frequent reviewer for funding agencies internationally and serves on the editorial boards of many IEEE and ACM journals. He currently serves as Chair of SIGSAC of ACM China. He is a Fellow of IEEE, a Fellow of ACM and a Clarivate Highly-Cited Researcher.
\end{IEEEbiography}